\documentclass[10pt,twocolumn,letterpaper]{article}

\usepackage[T1]{fontenc}

\usepackage{cvpr}
\usepackage{times}
\usepackage{epsfig}
\usepackage{graphicx}
\usepackage{amsmath}
\usepackage{amsthm}
\usepackage{amssymb}
\usepackage{multirow}
\usepackage{xcolor}
\usepackage{soul}
\usepackage{adjustbox}
\usepackage[numbers]{natbib}

\newcommand{\KL}[2]{\operatorname{KL}\big({\textstyle{#1}\,\|\,{#2}}\big)}
\newcommand{\D}{\mathcal{D}}
\newcommand{\Df}{{\mathcal{D}_f}}
\newcommand{\Dr}{{\mathcal{D}_r}}
\newcommand{\E}{\mathbb{E}}
\newcommand{\tr}{\operatorname{tr}}
\DeclareMathOperator{\Scrub}{S}

\usepackage[pagebackref=true,breaklinks=true,letterpaper=true,colorlinks,bookmarks=false]{hyperref}
\usepackage{cleveref}

\theoremstyle{definition}
\newtheorem{lemma}{Lemma}
\newtheorem{prop}{Proposition}
\newtheorem{corollary}{Corollary}
\newtheorem{definition}{Definition}

\newtheorem{example}{Example}

\cvprfinalcopy %

\def\cvprPaperID{6229} %

\ifcvprfinal\fi
\begin{document}

\title{Eternal Sunshine of the Spotless Net: Selective Forgetting in Deep Networks}

\author{Aditya Golatkar\\
UCLA\\
{\tt\small aditya29@cs.ucla.edu}
\and
Alessandro Achille%
\\
UCLA\\
{\tt\small achille@cs.ucla.edu}
\and 
Stefano Soatto\\
UCLA \\
{\tt\small soatto@ucla.edu}\\
}

\maketitle
\pagestyle{plain}

\begin{abstract}
We explore the problem of selectively forgetting a particular subset of the data used for training a deep neural network. While the effects of the data to be forgotten can be hidden from the output of the network, insights may still be gleaned by probing deep into its weights. We propose a method for ``scrubbing'' the weights clean of information about a particular set of training data. The method does not require retraining from scratch, nor access to the data originally used for training. Instead, the weights are modified so that any probing function of the weights is indistinguishable from the same function applied to the weights of a network trained without the data to be forgotten.
This condition is a generalized and weaker form of Differential Privacy.
Exploiting ideas related to the stability of stochastic gradient descent, we introduce an upper-bound on the amount of information remaining in the weights, which can be estimated efficiently even for deep neural networks.
\end{abstract}

\vspace{-0.6cm}
\section{Introduction}
Say you are the number `6' in the MNIST handwritten digit database. You are proud of having nurtured the development of convolutional neural networks and their many beneficial uses. But you are beginning to feel uncomfortable with the attention surrounding the new ``AI Revolution,'' and long to not be recognized everywhere you appear. You wish a service existed, like that offered by the firm Lacuna INC in the %
screenplay {\em The Eternal Sunshine of the Spotless Mind}, %
whereby you could submit your images to have your identity scrubbed clean from handwritten digit recognition systems.  Before you, the number `9' already demanded that digit recognition systems returned, instead of a ten-dimensional ``pre-softmax'' vector (meant to approximate the log-likelihood of an image containing a number from 0 to 9) a {\em nine}-dimensional vector that excluded the number `9'. So now, every image showing `9' yields an outcome at random between 0 and 8.
Is this enough? It could be that the system still contains {\em information} about the number '9,' and just suppresses it in the output. How do you know that the system has truly forgotten about you, even {\em inside the black box}? Is it possible to scrub the system so clean that it behaves as if it had never seen an image of you? Is it possible to do so without sabotaging information about other digits, who wish to continue enjoying their celebrity status?  In the next section we formalize these questions to address the problem of {\em selective forgetting in deep neural networks} (DNNs). Before doing so, we present a summary of our contributions in the context of related work.

\vspace{-0.1cm}
\subsection{Related Work}

Tampering with a learned model to achieve, or avoid, forgetting pertains to the general field of {\em life-long learning}. Specifically for the case of deep learning and representation learning, this topic has algorithmic, architectural and modeling ramifications, which we address in order.

\textbf{Differential privacy} \cite{dwork2014algorithmic} focuses on guaranteeing that the parameters of a trained model do not leak information about {\em any} particular individual. While this may be relevant in some applications, the condition is often too difficult to enforce in deep learning (although see \cite{abadi2016deep}), and not always necessary. 
It requires the possible distribution of weights, given the dataset, $P(w|\D)$ to remain almost unchanged after replacing a sample.
Our definition of selective forgetting can be seen as a generalization of differential privacy. In particular, we do not require that information about \emph{any} sample in the dataset is minimized, but rather about a particular subset $\Df$ selected by the user. Moreover, we can apply a ``scrubbing'' function $S(w)$ that can perturb the weights in order to remove information, so that $P(S(w)|\D)$, rather than $P(w|\D)$, needs to remain unchanged. This less restrictive setting allows us to train standard deep neural networks using stochastic gradient descent (SGD), while still being able to ensure forgetting.

Deep Neural Networks can memorize details about particular instances, rather than only shared characteristics \cite{zhang2016understanding,arpit2017closer}. This makes forgetting critical, as attackers can try to extract information from the weights of the model.
\textbf{Membership attacks} \cite{truex2019demystifying,hitaj2017deep,pyrgelis2017knock,hayes2019logan,song2017machine} attempt to determine whether a particular cohort of data was used for training, without any constructive indication on how to actively forget it. They relate to the ability of \textbf{recovering data from the model}
\cite{fredrikson2015model} which exploits the increased confidence of the model on the training data to reconstruct images used for training; 
\cite{micaelli2019zero} proposes a method for performing zero-shot knowledge distillation by adversarially generating a set of exciting images to train a student network.  
\cite{shintre2019verifying} proposes a definition of forgetting based on changes of the value of the loss function. We show that this is not meaningful forgetting, and in some cases it may lead to the (opposite) ``Streisand effect,'' where the sample to be forgotten is actually made more noticeable. 

\textbf{Stability of SGD.} In \cite{hardt2015train}, a bound is derived on the divergence of training path of  models trained with the same random seed (\ie, same initialization and sampling order) on datasets that differ by one sample (the ``stability'' of the training path). This can be considered as a measure of memorization of a sample and, thus,  used to bound the generalization error.
While these bounds are often loose, we introduce a novel bound on the residual information about a set of samples to be forgotten, which exploits ideas from both the stability bounds and the PAC-Bayes bounds \cite{mcallester2013pac}, which have been successful even for DNNs \cite{dziugaite2017computing}.

\textbf{Machine Unlearning} was first studied by \cite{cao2015towards} in the context of statistical query learning. \cite{bourtoule2019machine} proposed an unlearning method based on dataset sharding and training multiple models. \cite{ginart2019making} proposed an efficient data elimination algorithm for k-means clustering.  However, none of these methods can be applied for deep networks.
The term ``forgetting'' is also used frequently in life-long learning, but often with different connotations that in our work:   \textbf{Catastrophic forgetting}, where a network trained on a task rapidly loses accuracy on that task when fine-tuned for another. But while the network can  forget a {\em task,} the information on the {\em data} it used may still be accessible from the weights. Hence, even catastrophic forgetting does not satisfy our stronger definition. %
Interestingly, however, our proposed solution for forgetting relates to techniques used to {\em avoid} forgetting: \cite{kirkpatrick2017overcoming} suggests adding an $L_2$ regularizer using the Fisher Information Matrix (FIM) of the task. We use the FIM, restricted to the samples we wish to retain, to compute the optimal noise to destroy information, so that a cohort can be forgotten while maintaining good accuracy for the remaining samples. 
Part of our forgetting algorithm can be interpreted as performing ``optimal brain damage'' \cite{lecun1990optimal} in order to remove information from the weights if it is useful only or mainly to the class to be forgotten.

In this paper we talk about the weights of a network as containing ``information,'' even though we have {\em one} set of weights whereas information is commonly defined only for random variables. While this has caused some confusion in the literature, \cite{achille2019where} proposes a viable formalization of the notion, which is compatible with our framework. Thus, we will use the term ``information'' liberally even when talking about a particular dataset and set of weights.

In defining forgetting, we wish to be resistant to both ``black-box'' attacks, which only have access to the model output through some function (API), and ``white-box'' attacks, where the attacker can additionally access the model weights. Since at this point it is unclear how much information about a model can be recovered by looking only at its inputs and outputs, to avoid unforeseen weaknesses we characterize forgetting for the stronger case of white-box attacks, and derive bounds and defense mechanism for it.

\subsection{Contributions}

In summary, our contributions are, first, to propose a \textbf{definition of selective forgetting} for trained neural network models. It is not as simple as obfuscating the activations, and not as restrictive as Differential Privacy. 
Second, we propose a \textbf{scrubbing procedure} that removes information from the trained weights, without the need to access the original training data, nor to re-train the entire network.  We compare the scrubbed network to the gold-standard model(s) trained from scratch without any knowledge of the data to be forgotten. We also prove the optimality of this procedure in the quadratic case.
The approach is applicable to both the case where an entire class needs to be forgotten (e.g. the number `6') or multiple classes (e.g., all odd numbers), or a particular subset of samples within a class, while still maintaining output knowledge of that class. 
Our approach is applicable to networks pre-trained using standard loss functions, such as cross-entropy, unlike Differential Privacy methods that require the training to be conducted in a special manner.
Third, we introduce a \textbf{computable upper bound} to the amount of the retained information, which can be efficiently computed even for DNNs. We further characterize the optimal tradeoff with preserving complementary information. We illustrate the criteria using the MNIST and CIFAR-10 datasets, in addition to a new dataset called ``Lacuna.''

\subsection{Preliminaries and Notation}

Let $\D = \{x_i, y_i\}_{i=1}^N$ be a dataset of images $x_i$, each with an associated label $y_i \in \{1, \dots, K\}$ representing a class (or label, or identity). We assume that $(x_i, y_i) \sim P(x,y)$ are drawn from an unknown distribution $P$.

Let $\Df \subset \D$ be a subset of the data (cohort), whose information we want to remove (scrub) from a trained model, and let its complement $\Dr := \D_f^\complement$ be the data that we want to retain. The data to forget $\Df$ can be any subset of $\D$, but we are especially interested in the case where $\Df$ consists of all the data with a given label $k$ (that is, we want to completely forget about a class), or a subset of a class.

Let $\phi_w(\cdot) \in {\mathbb R}^K$ be a parametric function (model), for instance a DNN, with parameters $w$ (weights) trained using ${\cal D}$ so that the $k$-th component of the vector $\phi_w$ in response to an image $x$ approximates the optimal discriminant (log-posterior),  $\phi_w(x)_k \simeq  \log P(y=k|x)$, up to a normalizing constant. %

\subsection{Training algorithm and distribution of weights}
Given a dataset $\D$, we can train a model --- or equivalently a set of weights --- $w$ using some training algorithm $A$, that is $w = A(\D)$, where $A(\D)$ can be a stochastic function corresponding to a stochastic algorithm, for example stochastic gradient descent (SGD). Let $P(w|\D)$ denote the distribution of possible outputs of algorithm $A$, where $P(w|\D)$ will be a degenerate Dirac delta if $A$ is deterministic. The scrubbing function $S(w)$ --- introduced in the next section --- is also a stochastic function applied to the weights of a trained network. We denote by $P(S(w)|\D)$  the distribution of possible weights obtained  after training on the dataset $\D$ using algorithm $A$ and then applying the scrubbing function $S(w)$. Given two distributions $p(x)$ and $q(x)$, their Kullback-Leibler (KL) divergence is defined by $\KL{p(x)}{q(x)} := \E_{x\sim p(x)}\big[\log\big(p(x)/q(x)\big)\big]$.
The KL-divergence is always positive and can be thought of as a measure of similarity between distributions. In particular it is zero if and only if $p(x) = q(x)$.
Given two random variables $x$ and $y$, the amount of Shannon Mutual Information that $x$ has about $y$ is defined as $I(x; y) := \E_x \big[\KL{p(y|x)}{p(y)}\big]$.

\vspace{-0.15cm}
\section{Definition and Testing of Forgetting}

Let $\phi_w$ be a model trained on a dataset $\D = \Df \sqcup \Dr$ %
Then, a forgetting (or ``scrubbing'') procedure consists in applying a function $\Scrub(w; \Df)$ to the weights, with the goal of forgetting, that is to ensure that an ``attacker'' (algorithm) in possession of the model $\phi_w$ cannot  compute some ``readout function'' $f(w)$, to reconstruct information about $\Df$. 

It should be noted that one can always infer \textit{some} properties of $\Df$, even without having ever seen it. For example, if $\D$ consists of images of faces, we can infer that images in $\Df$ are likely to display two eyes, even without looking at the model $w$. What matters for forgetting is the amount of {\em additional} information $f(w)$ can extract from a cohort $\Df$ by exploiting the weights $w$, that could not have been inferred simply by its complement $\Dr$. 
This can be formalized as follows:
\begin{definition}
Given a readout function $f$, an optimal scrubbing function for $f$ is a function  $\Scrub(w; \Df)$ --- or $\Scrub(w)$, omitting the argument $\Df$ --- such that there is another function $\Scrub_0(w)$ that does not depend on $\Df$%
\footnote{
If $S_0$ could depend on $\Df$, we could take $S(w)=w$ to be the identity, and let $S_0(w)$ ignore $w$ and obtain new weights by training from scratch on $\D$ --- that is  $S_0(w) = w'$ with $w' \sim p(w|\D)$. This brings the KL to zero, but does not scrub any information, since $S(w)$ is the identity.
}
for which:
\begin{equation}
\label{eq:kl-forgetting}
    \KL{P(f(\Scrub(w))|\D)}{P(f(\Scrub_0(w))|\Dr)} = 0.
\end{equation}
\end{definition}
The function $S_0(w)$ in the definition  can be thought as a \emph{certificate} of forgetting, which shows that $S(w)$ is indistinguishable from a model that has never seen $\Df$. 
Satisfying the condition above is trivial by itself, \eg, by choosing $S(w)=S_0(w)=c$ to be constant. The point is to do so while retaining as much information as possible about $\D_r$, as we will see later when we introduce the Forgetting Lagrangian in \cref{eq:forgetting-lagrangian}. 
The formal connection between this definition and the amount of Shannon Information about $\Df$ that a readout function can extract is given by the following:
\begin{prop}
Let the forgetting set $\Df$ be a random variable, for instance, a random sampling of the data to forget. Let $Y$ be an attribute of interest that depends on $\Df$. Then,
\begin{align}
&I(Y; f(S(w))) \leq \nonumber\\ &\quad\quad\E_{D_f}[\KL{P(f(\Scrub(w))|\D)}{P(f(\Scrub_0(w))|\Dr)}].
\end{align}
\end{prop}

Yet another interpretation of \cref{eq:kl-forgetting} arises from noticing that, if that quantity is zero 
then, given the output of the readout function $f(w)$, we cannot predict with better-than-chance accuracy whether the model $w' = \Scrub(w)$ was trained with or without the data. In other words, after forgetting, membership attacks will fail.

In general, we may not know what readout function a potential attacker will use, and hence we want to be robust to every $f(w)$. The following lemma is useful to this effect:
\begin{lemma}
For any function $f(w)$ we have:
\begin{multline*}
\KL{P(f(\Scrub(w))|\D)}{P(f(\Scrub_0(w))|\Dr)}\\
\leq \KL{P(\Scrub(w)|\D)}{P(\Scrub_0(w)|\Dr)}. 
\end{multline*}
\end{lemma}
\noindent Therefore, we can focus on minimizing the quantity
\begin{equation}
\label{eq:kl-weights}
\KL{P(\Scrub(w)|\D)}{P(\Scrub_0(w)|\Dr)},
\end{equation}
which guarantees robustness to any readout function.
For the sake of concreteness, we now give a first simple example of a possible scrubbing procedure.
\begin{example}[Forgetting by adding noise]
Assume the weights $w$ of the model are bounded. Let $S(w) = S_0(w) = w + \sigma n$, where $n \sim {\cal N}(0, I)$, be the scrubbing procedure that adds noise sampled from a Gaussian distribution. Then, as the variance $\sigma$ increases, we achieve total forgetting:
\[
\KL{P(\Scrub(w)|\D)}{P(\Scrub_0(w)|\Dr)} \xrightarrow{\sigma \to \infty} 0.
\]
\end{example}
While adding noise with a large variance does indeed help forgetting, it throws away the baby along with the bath water, rendering the model useless. Instead, we want to forget as much as possible about a cohort while retaining the accuracy of the model. This can be formalized by minimizing the {\bf Forgetting Lagrangian}:
\begin{align}
\label{eq:forgetting-lagrangian}
\mathcal{L} &= \E_{S(w)} \big[L_{\Dr}(w)\big] \nonumber\\
&\quad\quad\quad+ \lambda \KL{P(\Scrub(w)|\D)}{P(\Scrub_0(w)|\Dr)},
\end{align}
where $L_\Dr(w)$ denotes the loss of the model $w$ on the retained data $\Dr$. Optimizing this first term is relatively easy. The problem is doing so while also minimizing the second (forgetting) term: For a DNN, the distribution $P(w|\D)$ of possible outcomes of the training process is complex making difficult the estimation of the KL divergence above, a problem we address in the next section. Nonetheless, the Forgetting Lagrangian, if optimized, captures the notion of {\bf selective forgetting} at the core of this work.

\subsection{Stability and Local Forgetting Bound}

Given a stochastic training algorithm $A(\D)$, we can make the dependency on the random seed $\epsilon$ explicit by writing $A(\D, \epsilon)$, where we assume that $A(\D, \epsilon)$ is now a deterministic function of the data and the random seed.
We now make the following assumptions: (1) the cohort to be forgotten, $\D_f$, is a small portion of the overall dataset $\D$, lest one is better-off re-training than forgetting, and (2) the training process $A(\D, \epsilon)$ is stable, \ie, if $\D$ and $\D'$ differ by a few samples, then the outcome of training $A(\D, \epsilon)$ is close to $A(\D', \epsilon)$.
Under stable training, we expect the two distributions $P(\Scrub(w)|\D)$ and $P(\Scrub_0(w)|\Dr)$ in \cref{eq:kl-weights} to be close, making forgetting easier.
Indeed, we now show how we can exploit the stability of the learning algorithm to bound the Forgetting Lagrangian.
\begin{prop}[Local Forgetting Bound]
\label{prop:local-forgetting-bound}
Let $A(\D, \epsilon)$ be a training algorithm with random seed $\epsilon \sim P(\epsilon)$. Notice that in this case $P(S(w)|\D) = \E_\epsilon[P(S(w)|\D, \epsilon)]$. We then have the bound:
\begin{multline*}
\KL{P(S(w)|\D)}{P(S_0(w)|\Dr)} \leq\\
 \E_\epsilon\Big[ \KL{P(S(w)|\D, \epsilon)}{P(S_0(w)|\Dr, \epsilon)} \Big]
\end{multline*}
\end{prop}
In the local forgetting bound we do not look at the global distribution of possible outcomes as the random seed varies, but only at the average of forgetting using a particular random seed. To see the value of this bound, consider the following example.
\begin{corollary}[Gaussian forgetting]
Consider the case where $S(w) = h(w) + n$ and $S_0(w) = w + n'$, where $n, n' \sim {\cal N}(0, \Sigma)$ is Gaussian noise and $h(w)$ is a deterministic function. Since for a fixed random seed $\epsilon$ the weights $w=A(\D, \epsilon)$ are a deterministic function of the data, we  have $P(S(w)|\D, \epsilon) = \mathcal{N}(h(A(\D, \epsilon)), \Sigma)$ and similarly $P(S_0(w)|\Dr, \epsilon) = \mathcal{N}(A(\Dr, \epsilon), \Sigma)$. Then, using the previous bound, we have:
\begin{multline}
\label{eq:gaussian-forgetting}
\KL{P(S(w)|\D)}{P(S_0(w)|\Dr)} \leq \\
\frac{1}{2} \E_\epsilon\Big[ (h(w) -w')^T\Sigma^{-1}(h(w) - w') \Big]
\end{multline}
where $w = A(\D, \epsilon)$ and $w' = A(\Dr, \epsilon)$.
\label{ex:gaussian-noise-forgetting}
\end{corollary}
That is, we can upper-bound the complex term $\KL{P(S(w)|\D)}{P(S_0(w)|\Dr)}$ with a much simpler one obtained by averaging the results of training  and scrubbing with different random seeds.

Moreover, this suggests three simple but general procedures to forget. Under the stability assumption, %
we can either (i) apply a function $h(w)$ that bring $w$ and $w'$ closer together (\ie, minimize $h(w) - w'$ in \cref{eq:gaussian-forgetting}), or (ii) add noise whose  covariance $\Sigma$ is high in the direction $h(w) - w'$, or (iii) both. Indeed, this will be the basis of our forgetting algorithm, which we describe next. 

\begin{figure}[t]
    \centering
    \includegraphics[width=.99\linewidth]{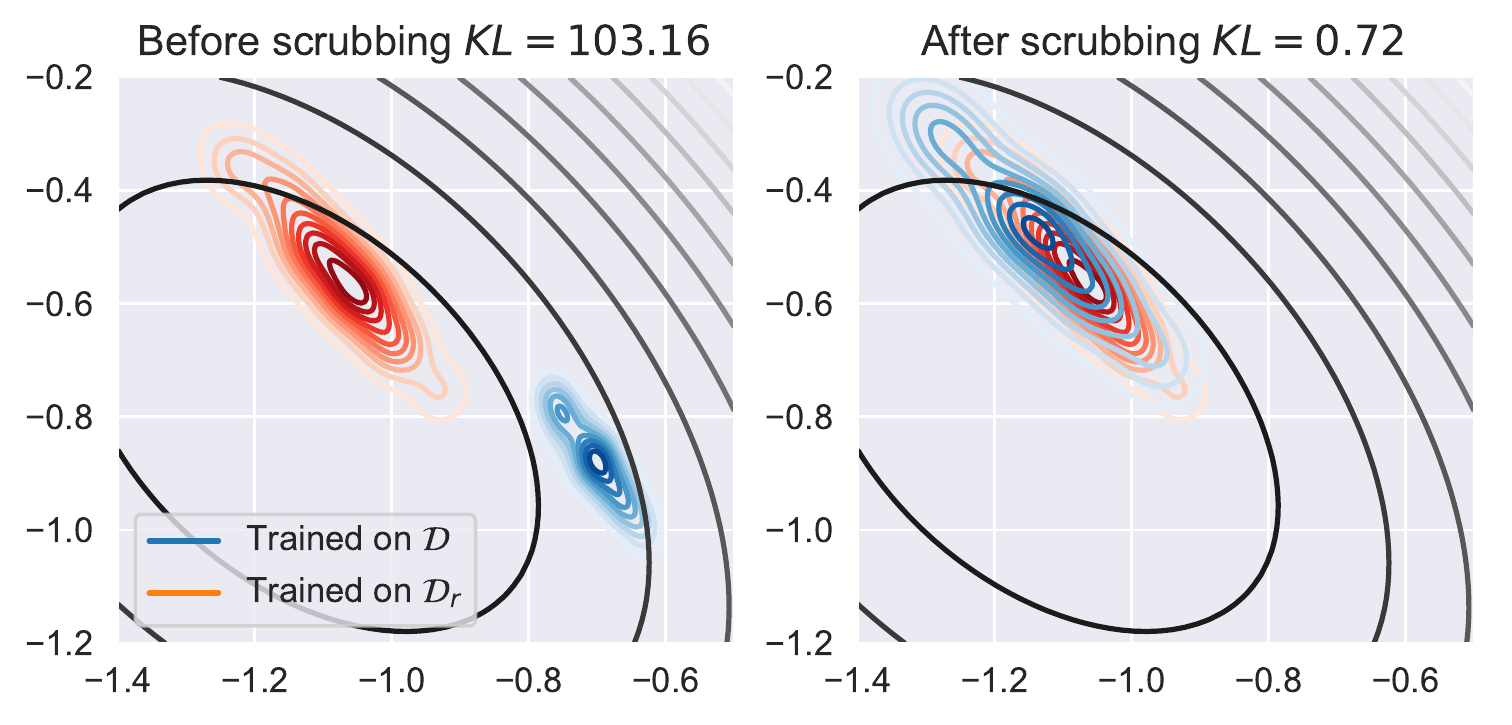}\\
    \includegraphics[width=.99\linewidth]{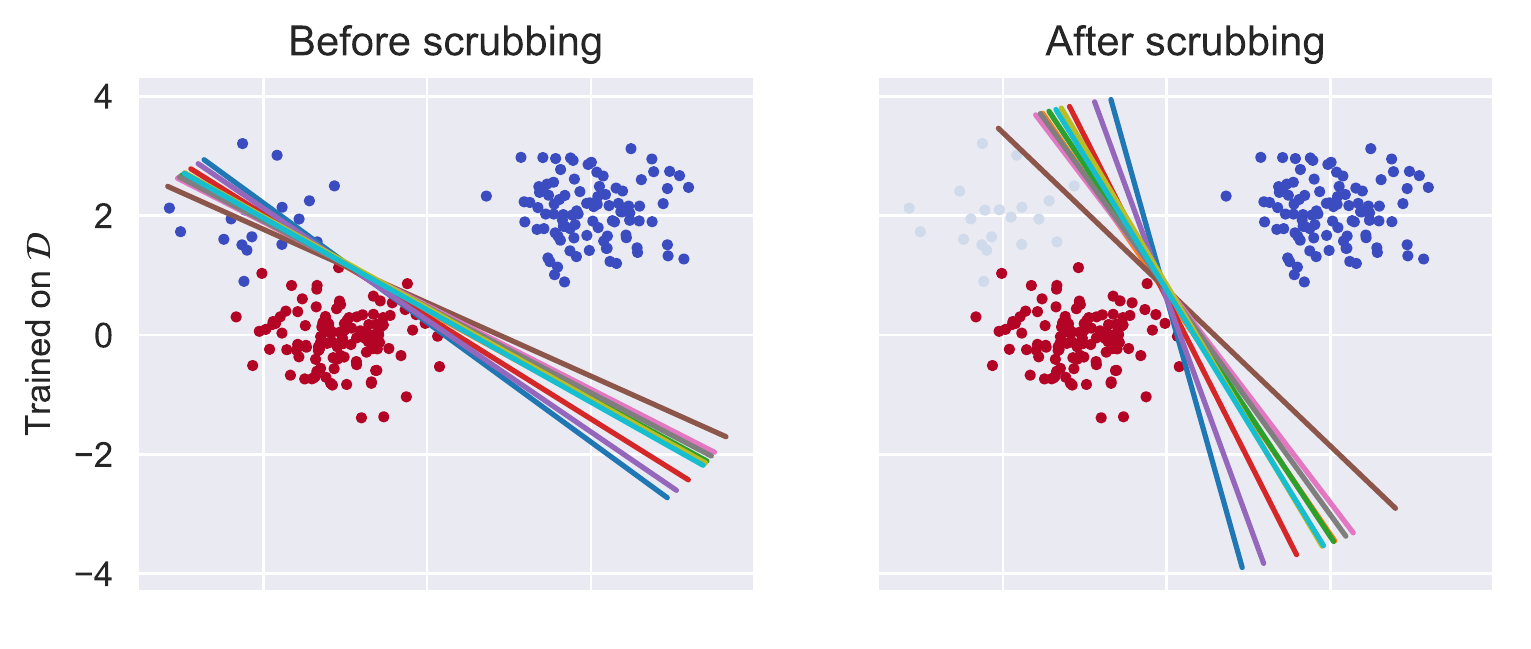}
    \caption{\textbf{(Top)} Distributions of weights $P(w|\D)$ and $P(w|\D_r)$ before and after the scrubbing procedure is applied to forget the samples $\D_f$. The scrubbing procedure makes the two distributions indistinguishable, thus preventing an attacker from extracting any information about $\D_f$. The KL divergence measures the maximum amount of information that an attacker can extract. After forgetting, less than 1 NAT of information about the cohort $\D_f$ is accessible. \textbf{(Bottom)} The effect of the scrubbing procedure on the distribution of possible classification boundaries obtained after training. After forgetting the subject on the top left blue cluster, the classification boundaries adjust as if she never existed, and the distribution mimics the one that would have been obtained by training from scratch without that data. }
    \label{fig:logistic-regression}
\end{figure}

\section{Optimal Quadratic Scrubbing Algorithm}

In this section, we derive an optimal scrubbing algorithm under a local quadratic approximation. We then validate the method empirically in complex real world problems where the assumptions are violated.
We start with strong assumptions, namely that the loss is quadratic and optimized in the limit of small learning rate, giving the continuous gradient descent optimization 
\[
A_t(\D, \epsilon) = w_0 - (I - e^{-\eta A t}) A^{-1} \nabla_w L_\D(w)|_{w=w_0},
\]
where $A=\nabla^2 L_\D(w)$ is the Hessian of the loss. We will relax these assumptions later.
\begin{prop}[Optimal quadratic scrubbing algorithm]
\label{prop:quadratic-scrubbing}
Let the loss be $L_{\D}(w) = L_{\Df}(w) + L_{\Dr}(w)$, and assume both $L_\D(w)$ and $L_\Dr(w)$ are quadratic. Assume that the optimization algorithm $A_t(\D, \epsilon)$ at time $t$ is given by the gradient flow on the loss with random initialization. Consider the scrubbing function
\[
h(w) = w + e^{-Bt}e^{At} d+ e^{-Bt}(d-d_r) - d_r,
\]
where $A = \nabla^2 L_\D(w)$, $B = \nabla^2 L_\Dr(w)$, $d = A^{-1} \nabla_w L_\D$ and $d_r = B^{-1} \nabla_w L_\Dr$.
Then, $h(w)$ is such that $h(A_t(\D, \epsilon)) = A_t(\D_r, \epsilon)$ for all random initializations $\epsilon$ and all times $t$. In particular,  $S(w) = h(w)$ scrubs the model clean of all information in $\D_f$:
\[
 \KL{P(S(w)|\D, \epsilon)}{P(w|\Dr, \epsilon)} = 0.
\]
Note that when $t\to \infty$, that is, after the optimization algorithm has converged, this reduces to the simple \emph{Newton update}:
\[
S_\infty(w) = w - B^{-1} \nabla L_\Dr(w).
\]
\end{prop}

\begin{figure}[t]
    \centering
    \includegraphics[width=.6\linewidth]{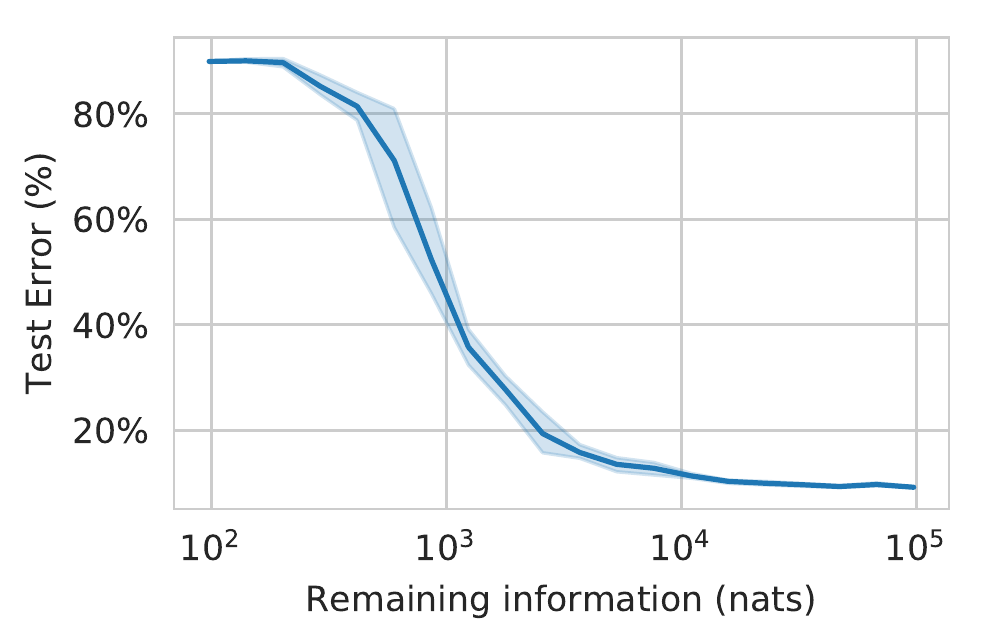}
    \caption{Trade-off between information remaining about the class to forget and test error, mediated by the parameter $\lambda$ in the Lagrangian: We can always forget more, but this comes at the cost of decreased accuracy.} 
    \label{fig:trade-off}
\end{figure}

\subsection{Robust Scrubbing}

\Cref{prop:quadratic-scrubbing} requires the loss to be quadratic, which is typically not the case. Even if it was, practical optimization proceeds in discrete steps, not as a gradient flow. To relax these assumptions, we exploit the remaining degree of freedom in the general scrubbing procedure introduced in \Cref{ex:gaussian-noise-forgetting}, which is the noise. 
\begin{prop}[Robust scrubbing procedure]
Assume that $h(w)$ is close to $w'$ up to some normally distributed error $h(w) - w' \sim N(0, \Sigma_h)$, and assume that $L_\Dr(w)$ is (locally) quadratic around $h(w)$. Then the optimal scrubbing procedure
in the form $S(w) = h(w) + n$, with $n \sim N(0, \Sigma)$, that minimizes the Forgetting Lagrangian \cref{eq:forgetting-lagrangian}
is obtained when $\Sigma B \Sigma = \lambda \Sigma_h$, where $B = \nabla^2 L_\Dr(w)$. In particular, if the error is isotropic, that is $\Sigma_h = \sigma_h^2 I$ is a multiple of the identity, we have $\Sigma = \sqrt{\lambda \sigma_h^2} B^{-1/2}$.
\end{prop}

Putting this together with the result in \Cref{prop:quadratic-scrubbing} gives us the following robust scrubbing procedure:
\begin{multline}
\label{eq:scrubbing-generic}
S_t(w) = w + e^{-Bt}e^{At} d+ e^{-Bt}(d-d_r) - d_r\\
+ (\lambda \sigma_h^2)^{\frac{1}{4}} B^{-1/4} n,    
\end{multline}
where $n \sim N(0, I)$ and $B$, $d$ and $d_r$ are as in \Cref{prop:quadratic-scrubbing}. In \Cref{fig:logistic-regression} we show the effect of the scrubbing procedure on a simple logistic regression problem (which is not quadratic) trained with SGD (which does not satisfy the gradient flow assumption). Nonetheless, the scrubbing procedure manages to bring the value of the KL divergence close to zero.
Finally, when $t \to \infty$ (\ie, the optimization is near convergence), this simplifies to the noisy Newton update which can be more readily applied:
\begin{equation}
\label{eq:scrubbing-newton}
\boxed{S_t(w) = w - B^{-1}\nabla L_\Dr(w) + (\lambda \sigma_h^2)^{\frac{1}{4}} B^{-1/4} \epsilon.}
\end{equation}

Here $\lambda$ is a hyperparameter that trades off residual information about the data to be forgotten, and accuracy on the data to be retained. The hyperparameter 
$\sigma_h$ reflect the error in approximating the SGD behavior with a continuous gradient flow.

\begin{figure}[t]
    \centering
    \includegraphics[width=.9\linewidth]{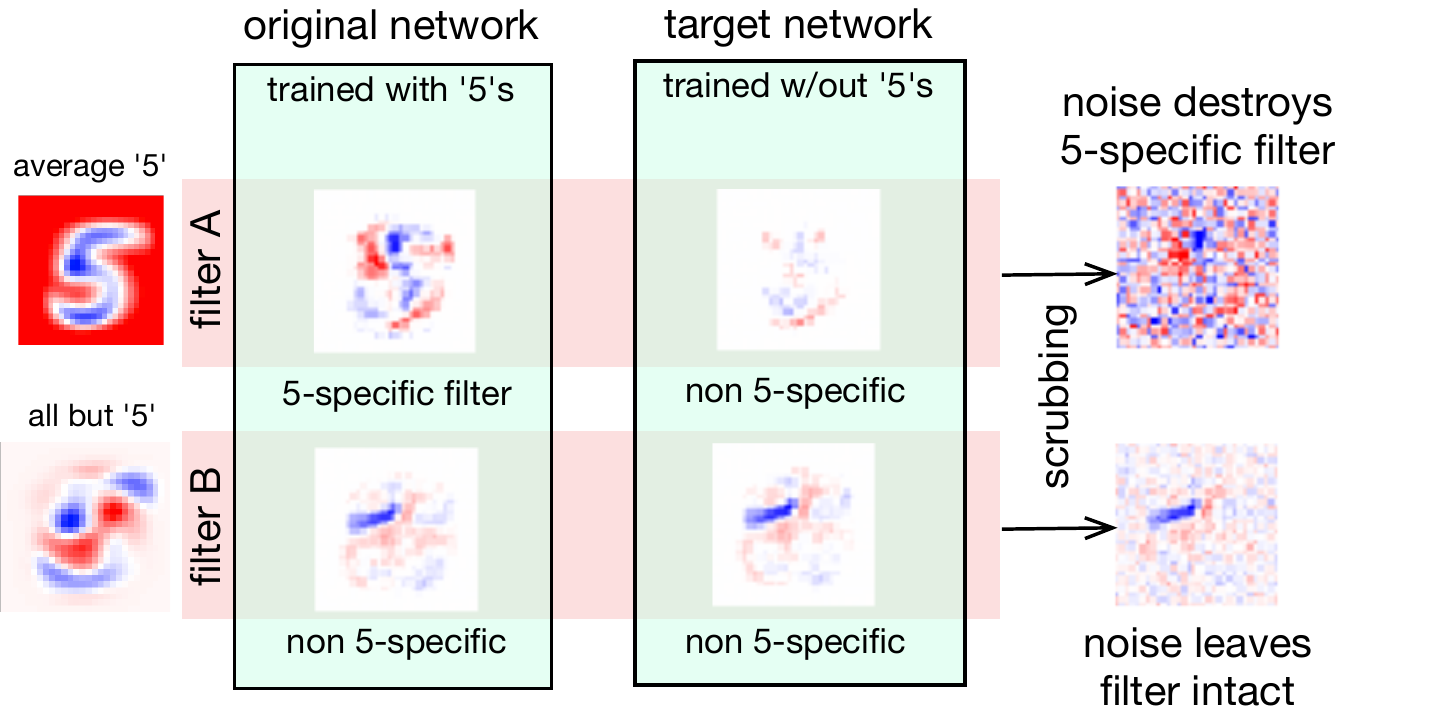}
    \caption{Filters of a network trained with the same random seed, with and without 5's. Some filters specialize to be 5-specific (filter A), and differ between the two networks, while others are not 5-specific (filter B), and remain identical.
    The scrubbing procedure brings original and target network closer by destroying 5-specific filters, effectively removing information about 5's.}
    \label{fig:mnist-filters}
\end{figure}

\subsection{Forgetting using a subset of the data}

Once a model is trained, a request to forget $\Df$ may be initiated by providing that cohort, as in the fictional service of Lacuna INC, but in general one may no longer have available the remainder of the dataset used for training, $\Dr$. However,
assuming we are at a minimum of $L_\D(w)$, we have $\nabla L_\D(w) = 0$. Hence, we can rewrite $\nabla L_{\Dr}(w) = - \nabla L_{\Df}(w)$ and $\nabla^2 L_{\Df}(w) = \nabla^2 L_{\D}(w) - \nabla^2 L_{\Dr}(w)$. Using these identities, instead of recomputing the gradients and Hessian on the whole dataset, we can simply use those computed on the cohort to be forgotten, provided we cached the Hessian $\nabla^2 L_{\D}(w)$ we obtained at the end of the training on the original dataset $\D$. Note that this is not a requirement, although recommended in case the data to be remembered is no longer available.

\begin{table*}[t]
\centering

\vspace{.5em}
\newcommand{\std}[1]{\color{black!70}{$\pm$#1}}
\resizebox{2\columnwidth}{!}{
\begin{adjustbox}{center}
{\small
\begin{tabular}{p{0.075\linewidth}|p{0.14\linewidth}|p{0.075\linewidth}|p{0.075\linewidth}|p{0.074\linewidth}p{0.073\linewidth}p{0.074\linewidth}p{0.074\linewidth}|p{0.089\linewidth}p{0.082\linewidth}}
\hline
  &  Metrics & \small{\textbf{Original model}} & \small{\textbf{Retrain} (target)} & \small{\textbf{Finetune}} & \small{\textbf{Neg.~Grad.}} & \small{\textbf{Rand.~Lbls.}}  & \small{\textbf{Hiding}}  & \small{\textbf{Fisher} (ours)} & \small{\textbf{Variational} (ours)}  \\
\hline
\footnotesize{Lacuna-10} & Error on $\mathcal{D}_\text{test}$ \scriptsize{(\%)} & 
\footnotesize{10.2} \std{0.5} & 
\footnotesize{10.3} \std{0.4} & 
\footnotesize{10.2} \std{0.6} & 
\footnotesize{10.0} \std{0.4}  & 
\footnotesize{12.0} \std{0.2} & 
\footnotesize{18.2} \std{0.4} & 
\footnotesize{14.5} \std{1.6} & 
\footnotesize{13.7} \std{1.0}\\
\scriptsize{Scrub 100} & Error on $\mathcal{D}_f$ \scriptsize{(\%)} & \footnotesize{0.0} \std{0.0} & 
\footnotesize{15.3} \std{0.6} & 
\footnotesize{0.0} \std{0.0} & 
\footnotesize{0.0} \std{0.0} & 
\footnotesize{6.0} \std{3.6} & 
\footnotesize{100} \std{0.0} & 
\footnotesize{8.0} \std{2.7} & 
\footnotesize{8.0} \std{3.6}\\
\scriptsize{All-CNN} & Error on  $\mathcal{D}_r$ \scriptsize{(\%)} & 
\footnotesize{0.0} \std{0.0} & 
\footnotesize{0.0} \std{0.0} & 
\footnotesize{0.0} \std{0.0} & 
\footnotesize{0.0} \std{0.0} & 
\footnotesize{0.1} \std{0.1} & 
\footnotesize{6.5} \std{0.0} & 
\footnotesize{4.8} \std{2.8} & 
\footnotesize{4.8} \std{2.4}\\
\scriptsize{} & Info-bound \scriptsize{(kNATs)} &    &  &  &   &  &  & \footnotesize{3.3} \std{1.1} &  \footnotesize{3.0} \std{0.5}\\
\hline
\footnotesize{Lacuna-10} & Error on $\mathcal{D}_\text{test}$ \scriptsize{(\%)} & 
\footnotesize{10.2} \std{0.5} & 
\footnotesize{18.4} \std{0.6} & 
\footnotesize{10.0} \std{0.6} & 
\footnotesize{18.4} \std{0.6}  & 
\footnotesize{18.8} \std{0.6} & 
\footnotesize{18.2} \std{0.4} & 
\footnotesize{21.0} \std{1.3} & 
\footnotesize{20.9} \std{0.4}\\
\scriptsize{Forget class} & Error on $\mathcal{D}_f$ \scriptsize{(\%)} & 
\footnotesize{0.0} \std{0.0} & 
\footnotesize{100} \std{0.0} & 
\footnotesize{0.0} \std{0.0} & 
\footnotesize{100} \std{0.2}  & 
\footnotesize{90.2} \std{1.5} & 
\footnotesize{100.0} \std{0.0} & 
\footnotesize{100.0} \std{0.0} & 
\footnotesize{100.0} \std{0.0}\\

\scriptsize{All-CNN} & Error on  $\mathcal{D}_r$ \scriptsize{(\%)} & 
\footnotesize{0.0} \std{0.0} & 
\footnotesize{0.0} \std{0.0} & 
\footnotesize{0.0} \std{0.0} & 
\footnotesize{0.0} \std{0.0}  & 
\footnotesize{0.0} \std{0.0} & 
\footnotesize{0.0} \std{0.0} & 
\footnotesize{3.3} \std{2.3} & 
\footnotesize{2.8} \std{1.4}\\
& Info-bound \scriptsize{(kNATs)}&  & &  & &  &  &  
\footnotesize{13.2} \std{2.8} &  \footnotesize{12.0} \std{2.9}\\
\hline
\footnotesize{CIFAR-10} & Error on $\mathcal{D}_\text{test}$ \scriptsize{(\%)} & 
\footnotesize{14.4} \std{0.6} & 
\footnotesize{14.6} \std{0.7} & 
\footnotesize{13.5} \std{0.1} & 
\footnotesize{13.4} \std{0.1} & 
\footnotesize{13.8} \std{0.1} & 
\footnotesize{21.0} \std{0.5} & 
\footnotesize{19.8} \std{2.8} & 
\footnotesize{20.9} \std{4.8}\\
\scriptsize{Scrub 100} & Error on $\mathcal{D}_f$ \scriptsize(\%)& 
\footnotesize{0.0} \std{0.0} & 
\footnotesize{19.3} \std{4.5} & 
\footnotesize{0.0} \std{0.0} & 
\footnotesize{0.0} \std{0.0} & 
\footnotesize{0.0} \std{0.0} & 
\footnotesize{100.0} \std{0.0} & 
\footnotesize{23.3} \std{2.1} & 
\footnotesize{6.3} \std{2.5}\\
\scriptsize{All-CNN} & Error on  $\mathcal{D}_r$ \scriptsize(\%)& 
\footnotesize{0.0} \std{0.0} & 
\footnotesize{0.0} \std{0.0} & 
\footnotesize{0.0} \std{0.0} & 
\footnotesize{0.0} \std{0.0} & 
\footnotesize{0.0} \std{0.0} & 
\footnotesize{9.9} \std{0.1} & 
\footnotesize{8.0} \std{4.3} & 
\footnotesize{8.8} \std{5.2}\\
\scriptsize{} & Info-bound \scriptsize{(kNATs)} &  &  &  &   &  &  & \footnotesize{33.4} \std{16.7} &  \footnotesize{21.6} \std{5.2}\\
\hline
\footnotesize{CIFAR-10} & Error on $\mathcal{D}_\text{test}$ \scriptsize{(\%)} & 
\footnotesize{14.4} \std{0.7} & 
\footnotesize{21.1} \std{0.6} & 
\footnotesize{14.3} \std{0.1} & 
\footnotesize{20.2} \std{0.1} & 
\footnotesize{20.7} \std{0.4} & 
\footnotesize{21.0} \std{0.5} & 
\footnotesize{23.7} \std{0.9} & 
\footnotesize{22.8} \std{0.3}\\
\scriptsize{Forget class} & Error on $\mathcal{D}_f$ \scriptsize{(\%)} & 
\footnotesize{0.0} \std{0.0} & 
\footnotesize{100} \std{0.0} & 
\footnotesize{10.0} \std{0.4} & 
\footnotesize{100} \std{0.2}  & 
\footnotesize{88.1} \std{4.6} & 
\footnotesize{100.0} \std{0.0} & 
\footnotesize{100.0} \std{0.0} & 
\footnotesize{100.0} \std{0.0}\\
\scriptsize{All-CNN} & Error on  $\mathcal{D}_r$ \scriptsize{(\%)} & 
\footnotesize{0.0} \std{0.0} & 
\footnotesize{0.0} \std{0.0} & 
\footnotesize{0.0} \std{0.0} & 
\footnotesize{0.0} \std{0.0}  & 
\footnotesize{0.0} \std{0.0} & 
\footnotesize{0.0} \std{0.0} & 
\footnotesize{2.6} \std{1.8} & 
\footnotesize{2.3} \std{0.7}\\
& Info-bound \scriptsize{(kNATs)}& & &  & &  &  &  
\footnotesize{458.1} \std{172.2} &  \footnotesize{371.5} \std{51.3}\\
\hline
\hline
\end{tabular}
}
\end{adjustbox}
}
\caption{
\label{table:results}
\textbf{Original model} is the model trained on all data $\D = \Df \sqcup \Dr$. The forgetting algorithm should scrub information from its weights. \textbf{Retrain} denotes the model obtained by retraining from scratch on $\D_r$, without knowledge of $\D_f$. The metric values in the Retrain column is the optimal value which every other scrubbing procedure should attempt to match. We consider the following forgetting procedures: \textbf{Fine-tune} denotes fine-tuning the model on $\D_r$. \textbf{Negative Gradient} (Neg. Grad.) denotes fine-tuning on $\D_f$ by moving in the direction of increasing loss. \textbf{Random Label} (Rnd. Lbls.) denotes replacing the labels of the class with random labels and then fine-tuning on all $\D$. \textbf{Hiding} denotes simply removing the class from the final classification layer. \textbf{Fisher} and \textbf{Variational} are our proposed methods, which add noise to the weights to destroy information about $\D_f$ following the Forgetting Lagrangian. We benchmark these methods using several readout functions: errors on $\Df$ and $\Dr$ after scrubbing, time to retrain on the forgotten samples after scrubbing, distribution of the model entropy.
In all cases, the read-out of the scrubbed model should be closer to the target retrained model than to the original.
Note that our methods also provide an upper-bound to the amount of information remaining. We report mean/std over 3 random seeds.
}
\end{table*}

\subsection{Hessian approximation and Fisher Information}

In practice, the Hessian is too expensive to compute for a DNN. In general, we cannot even ensure it is positive definite. To address both issues, we use the  Levenberg-Marquardt semi-positive-definite approximation:
\begin{equation}
\label{eq:fisher}
\nabla^2 L_{\D}(w) \simeq \E_{x \sim \D,y\sim p(y|x)}[\nabla_w \log p_w(y|x) \nabla_w \log p_w(y|x)^T].
\end{equation}
This approximation of the Hessian coincides with the Fisher Information Matrix (FIM) \cite{martens2014new}, which opens the door to information-theoretic interpretations of the scrubbing procedure. Moreover, this approximation is exact for some problems, such as linear (logistic) regression.

\section{Deep Network Scrubbing} 

Finally, we now discuss how to robustly apply the forgetting procedure \cref{eq:scrubbing-newton} to deep networks. We present two variants. The first uses the FIM of the network. However, since this depends on the network gradients, it may not be robust when the loss landscape is highly irregular. To solve this, we present a more robust method that attempts to minimize directly the Forgetting Lagrangian \cref{eq:forgetting-lagrangian} through a variational optimization procedure.

\vspace{.5em}
\noindent\textbf{Fisher forgetting:}
As mentioned in \cref{eq:fisher}, we approximate the Hessian with the Fisher Information Matrix. Since the FIM is too large to store in memory, we can compute its diagonal, or a better Kronecker-factorized approximation \cite{martens2015optimizing}. In our experiments, we find that  the diagonal is not a good enough approximation of $B$ for a full Newton step $h(w) = w - B^{-1} \nabla L_\Dr(w)$ in \cref{eq:scrubbing-newton}.
However, the diagonal is still a good approximation for the purpose of adding noise. Therefore, we simplify the procedure and take $h(w) = w$, while we still use the approximation of the FIM as the covariance of the noise.  This results in the simplified scrubbing procedure:
\[
\Scrub(w) = w + (\lambda \sigma_h^2)^{\frac{1}{4}} F^{-1/4},
\]
where $F$ is the FIM (eq. \ref{eq:fisher}) computed at the point $w$ for the dataset $\Dr$.
Here $\lambda$ is a hyper-parameter that trades off  forgetting with the increase in error, as shown in \Cref{fig:trade-off}. 
Notice that, since $h(w)=w$, instead of a Newton step, this procedure relies on $w$ and $w'$ already being close, which hinges on the stability of SGD.
This procedure may be interpreted as adding noise to destroy the weights that may have been informative about  $\Df$ but not $\Dr$ (\Cref{fig:mnist-filters}).

\begin{figure*}[t]
    \centering
    \includegraphics[width=.9\linewidth]{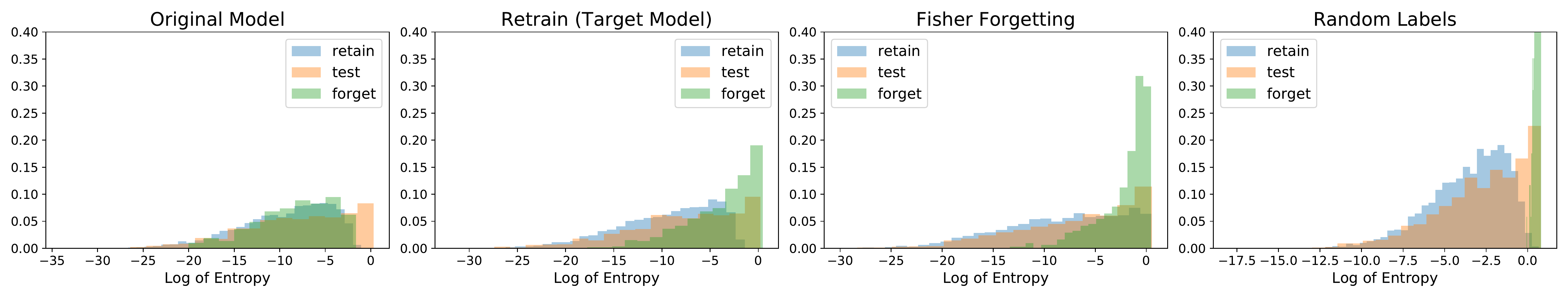}
    \caption{{\bf Streisand Effect:} Distribution of the entropy of model output (confidence) on: the retain set $\D_r$, the forget set $\D_f$, and the test set. The \textbf{original model} has seen $\D_f$, and its prediction on it are very confident (matching the confidence on the train data). On the other hand, a model \textbf{re-trained} without seeing $\D_f$ has a lower confidence $\D_f$. After applying our scrubbing procedures (\textbf{Fisher} and \textbf{Variational}) to the original model, the confidence matches more closely the one we would have expected for a model that has never seen the data (column 3 is more similar to 2 than 1). For an incorrect method of forgetting, like training with random labels, we observe that the entropy of the forgotten samples is very degenerate and different from what we would have expected if the model had actually never seen those samples (it is concentrated only around chance level prediction entropy).
    That is, attempting to remove information about a particular cohort using this method, may actually end up providing more information about the cohort than the original model.
    }  

    \label{fig:streisand}
\end{figure*}

\vspace{.5em}
\noindent\textbf{Variational forgetting:}
Rather than using the FIM, we may optimize for the noise in the Forgetting Lagrangian in  \cref{eq:forgetting-lagrangian}: Not knowing the optimal direction $w - w'$ along which to add noise (see \Cref{ex:gaussian-noise-forgetting}), we may add the maximum amount of noise in all directions, while keeping the increase in the loss to a minimum. Formally, we minimize the proxy Lagrangian:
\[
\mathcal{L}(\Sigma) = \E_{n \sim N(0, \Sigma)} \big[L_\Dr(w+n)\big] - \lambda \log |\Sigma|.
\]
The optimal $\Sigma$ may be seen as the FIM computed over a smoothed landscape. Since the noise is Gaussian, $\mathcal{L}(\Sigma)$ can be optimized using the local reparametrization trick \cite{kingma2015variational}.

\section{Experiments}

We report experiments on MNIST, CIFAR10 \cite{krizhevsky2009learning}, Lacuna-10 and Lacuna-100, which we introduce and consist respectively of faces of 10 and 100 different celebrities from VGGFaces2 \cite{Cao18} (see Appendix for details). On both CIFAR-10 and Lacuna-10 we choose to forget either an entire class, or a hundred images of the class.

For images (Lacuna-10 and CIFAR10), we use  a small All-CNN (reducing the number of layers) \cite{springenberg2014striving}, to which we add batch normalization before each non-linearity. 
We pre-train on Lacuna-100/CIFAR-100 for 15 epochs using SGD with fixed learning rate of 0.1, momentum 0.9 and weight decay 0.0005. We  fine-tune on Lacuna-10/CIFAR-10 with learning rate 0.01. To simplify the analysis, during fine-tuning we do not update the running mean and variance of batch normalization, and rather reuse the pre-trained ones.

\subsection{Linear logistic regression}

First, to validate the theory, we test the scrubbing procedure in \cref{eq:scrubbing-generic} on logistic regression, where the task is to forget data points belonging to one of two clusters comprising the class (see \Cref{fig:logistic-regression}). We train using a uniform random initialization for the weights and SGD with batch size 10, with early stopping after 10 epochs. Since the problem is low-dimensional, we easily approximate the distribution $p(w|\D)$ and $p(w|\Dr)$ by training 100 times with different random seeds.
As can be seen in \Cref{fig:logistic-regression}, the scrubbing procedure is able to align the two distributions with near perfect overlap, therefore preventing an attacker form extracting any information about the forgotten cluster. Notice also that, since we use early stopping, the algorithm had not yet converged, and exploiting the time dependency in \cref{eq:scrubbing-generic} rather than using the simpler \cref{eq:scrubbing-newton} is critical.

\subsection{Baseline forgetting methods}
Together with our proposed methods, we experiment with four other baselines which may intuitively provide some degree of forgetting. (i) \textbf{Fine-tune}: we fine-tune the model on the remaining data $\D_{r}$ using a slightly large learning rate. This is akin to catastrophic forgetting, where fine-tuning without $\D_f$ may make the model forget the original solution to $\D_f$ (more so because of the larger learning rate). (ii) \textbf{Negative Gradient}: we fine-tune on $\D$ by moving in the direction of increasing loss for samples in $\D_{f}$, which is equivalent to using a negative gradient for the samples to forget. This aims to damage features predicting $\D_f$ correctly. (iii) \textbf{Random Labels}: fine-tune the model on $\D$ by randomly resampling labels corresponding to images belonging to $\Df$, so that those samples will get a random gradient. (iv) \textbf{Hiding}: we simply remove the row corresponding to the class to forget from the final classification layer of the DNN.

\subsection{Readout functions used}

Unlike our methods, these baselines do not come with an upper bound on the quantity of remaining information. It is therefore unclear how much information is removed. For this reason, we introduce the following \textit{read-out functions}, which may be used to gauge how much information they were able destroy:
 (i) \textbf{Error on the test set} $\D_\text{test}$ (ideally small), (ii) \textbf{Error on the cohort to be forgotten} $\Df$ (ideally the same as a model trained without seeing $\Df$), (iii) \textbf{Error on the residual  $\Dr$} (ideally small), (iv) \textbf{Re-learn time} (in epochs) time to retrain the scrubbed model on the forgotten data (measured by the time for the loss to reach a fixed threshold, ideally slow). If a scrubbed model can quickly recover a good accuracy, information about that cohort is likely still present in the weights. (v) \textbf{Model confidence:} We plot the distribution of model confidence (entropy of the output prediction) on the retain set $\Dr$, forget set $\Df$ and the test set (should look similar to the confidence of a model that has never seen the data).
(vi) \textbf{Information bound}: For our methods, we compute the information upper-bound  about the cohort to be forgotten in NATs using \Cref{prop:local-forgetting-bound}.

\begin{figure}[t]
    \centering
    \includegraphics[width=.90\linewidth]{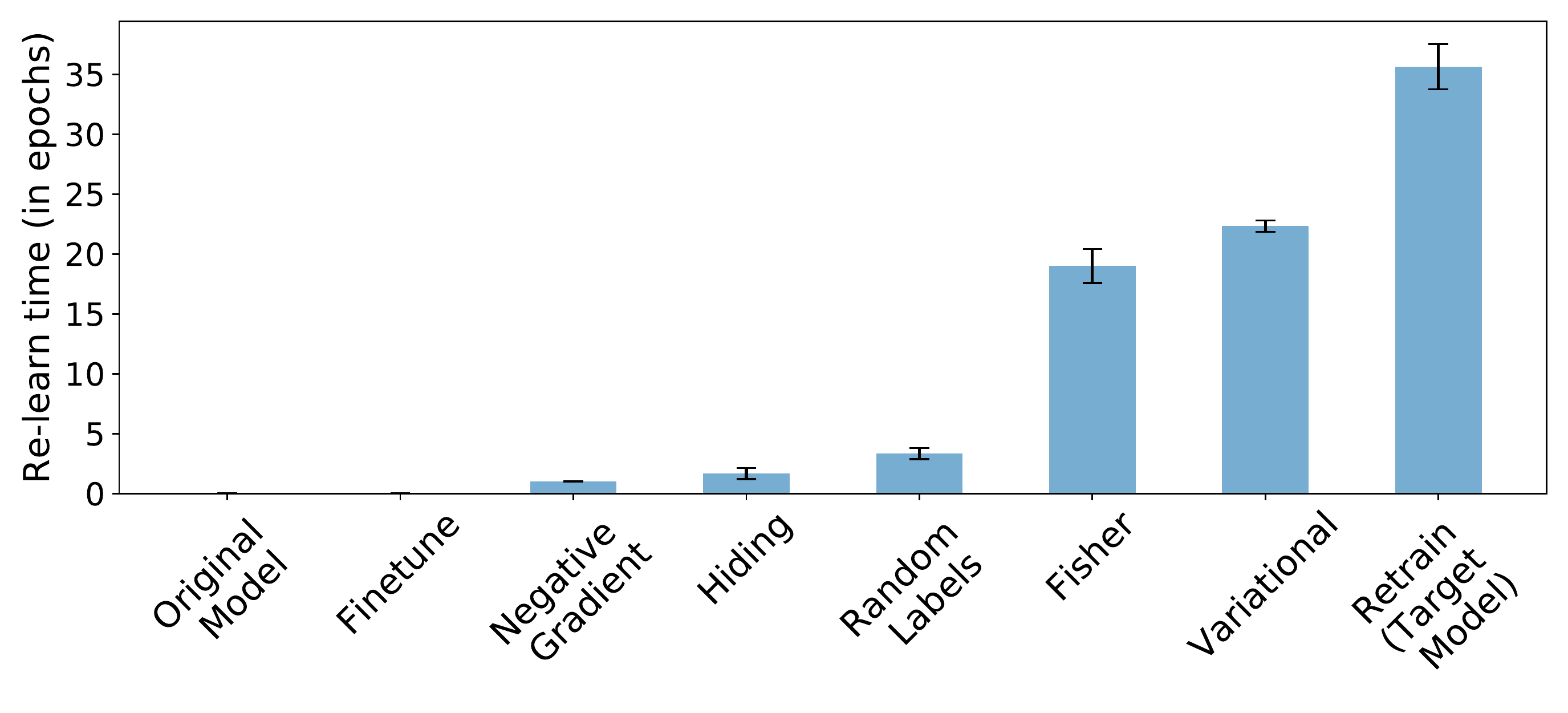}
    \caption{\textbf{Re-learn time} (in epochs) for various forgetting methods. All the baselines method can quickly recover perfect performance on $\Df$, suggesting that they do not actually scrub information from the weights. On the other hand, the relearn time for our methods is higher, and closer to the one of a model that has never seen the data, suggesting that they remove more information.
    } 
    \label{fig:lacuna-relearn}
\end{figure}

\subsection{Results}

First, in \Cref{table:results} we show the results of scrubbing  $\Df$ from model trained on all the data. We test both the case we want to forget only a subset of 100-images from the class, and when we want to forget a whole identity. We test on CIFAR-10 and Lacuna-10 with a network pretrained on CIFAR-100 and Lacuna-100 respectively.

Retrain denotes the gold standard which every scrubbing procedure should attempt to match for the error readout function. From the case where we want to scrub a subset of a class (first and third row of Retrain) it is clear that scrubbing does not mean merely achieving 100\% error on $\Df$. In fact, the reference Retrain has 15.3\% and 19.3\% error respectively on $\Df$ and not 100\%. Rather it means removing the information from the weights so that it behaves identically to a re-trained model. Forgetting by fine-tuning on $\Dr$, performs poorly on the error readout function (error on $\Df$ and $\Dr$), suggesting that using catastrophic forgetting is the not the correct solution to selective forgetting.

The Negative Gradient and Random Labels methods perform well on the error readout function, however, when we use the re-learn time as a readout function (\Cref{fig:lacuna-relearn}) it becomes clear that very little information is actually removed, as the model relearn $\Df$ very quickly. This suggests that merely scrubbing the activations by hiding or changing some output is not sufficient for selective forgetting; rather, information needs to be removed from the weights as anticipated.
Moreover, applying an incorrect scrubbing procedure may make the images to forget \textit{more} noticeable to an attacker (Streisand effect), as we can see by from the confidence values in \Cref{fig:streisand}.
The ease of forgetting a learnt cohort also depends on its size. In particular, in \Cref{fig:info-vs-samples} we observe that, for a fixed value of $\lambda$ in \cref{eq:kl-forgetting}, the upper-bound on the information retained by the model after scrubbing increases with the size of the cohort to forget.

\begin{figure}[b]
    \centering
    \includegraphics[width=0.75\linewidth]{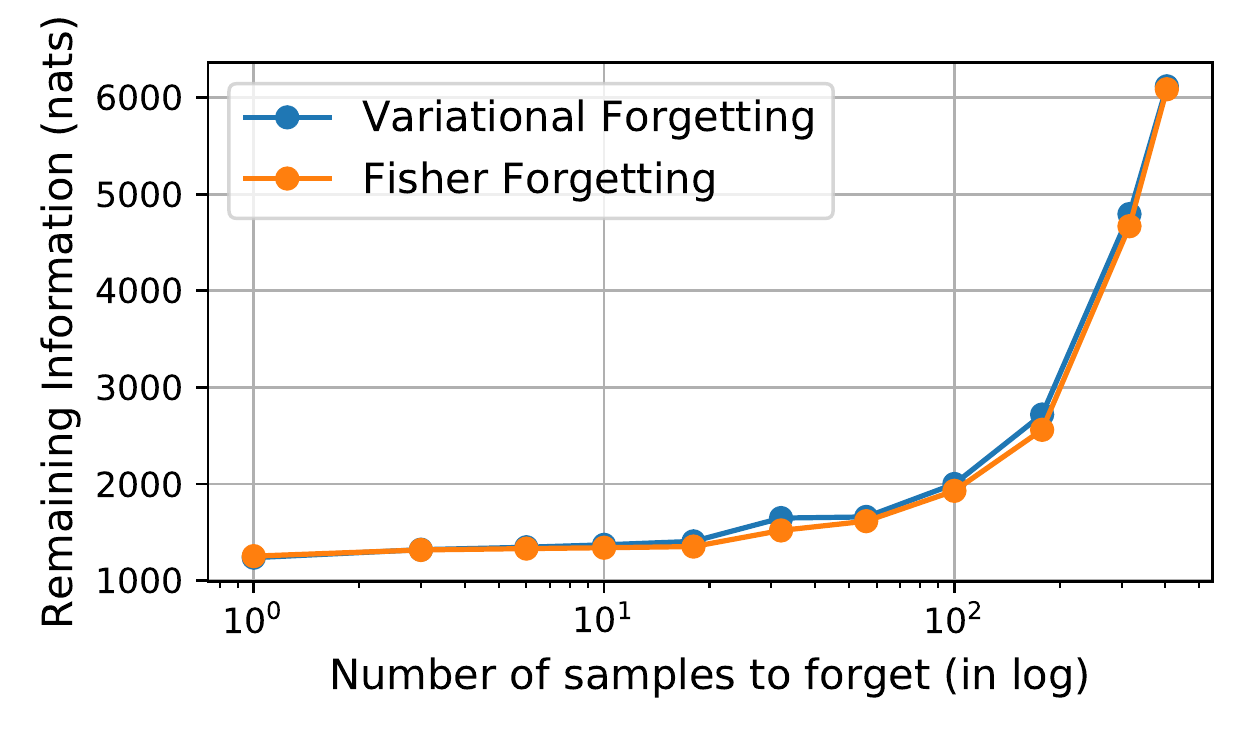}
    \caption{\textbf{Difficulty of forgetting increases with cohort size.} For a fixed $\lambda$ (forgetting parameter), we plot the amount of information remaining after scrubbing as a function of the cohort size ($|\Df|$).}
    \label{fig:info-vs-samples}
\end{figure}

\vspace{-0.47cm}
\section{Discussion}

Our approach is rooted in the %
connection between Differential Privacy (which our framework generalizes) and the stability of SGD.
Forgetting is also intrinsically connected with  information: Forgetting may also be seen as minimizing an upper-bound on the amount of information that the weights contain about $\Df$ \cite{achille2019where} and that an attacker may extract about that particular cohort $\Df$ using some readout function $f$. We have studied this problem from the point of view of Shannon Information, which allows for an easy formalization. However, it also has the drawback of considering the worst case of an attacker that has full knowledge of the training procedure and can use arbitrarily complex readout functions which may, for example, simulate all possible trainings of the network to extract the result.  Characterizing forgetting with respect to a viable subset of realistic readout functions $f(w)$ is a promising area of research.
We also exploit stability of the training algorithm after pretraining. Forgetting without the pretraining assumption is an interesting challenge, as it has been observed that slight perturbation of the initial critical learning period can lead to large difference in the final solution \cite{achille2017critical,golatkar2019time}.

\vspace{.5em}
\noindent\textbf{Acknowledgements:} We would like to thank the anonymous reviewers for their feedback and suggestions. This work is supported by ARO W911NF-17-1-0304, ONR N00014-17-1-2072, ONR N00014-19-1-2229, ONR N00014-19-1-2066.

{\small
\bibliographystyle{ieee_fullname}
\bibliography{references}
}
\clearpage
\appendix
\centerline{\Large \bf Supplementary Material}
\begin{figure*}[t]
    \centering
    \includegraphics[width=.55\linewidth]{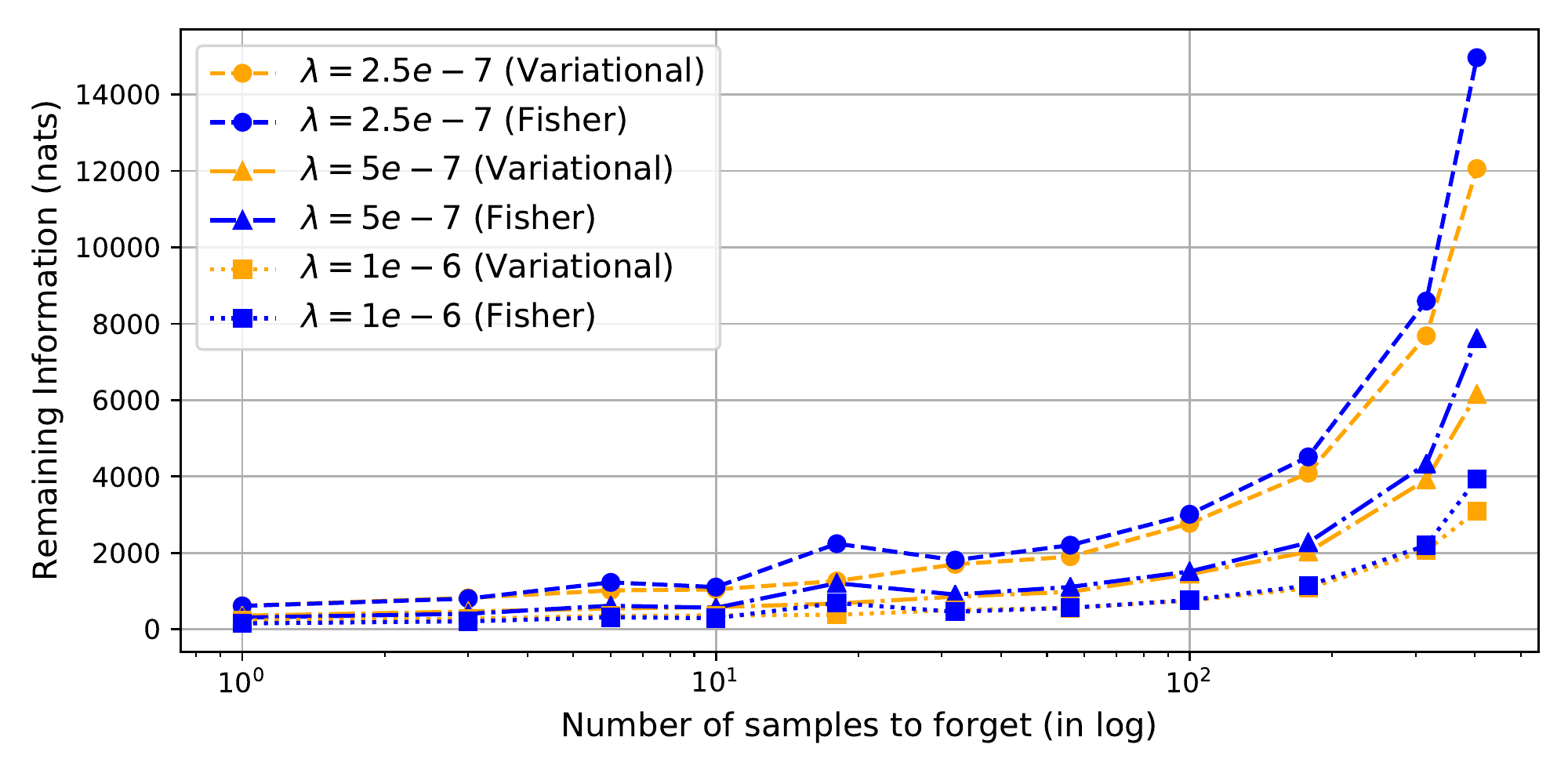}
    \caption{\textbf{Difficulty of forgetting increases with cohort size (for ResNet)} We plot the upper-bound on the remaining information (i.e. the information the model contains about the cohort to forget after scrubbing) as a function of the number of samples to forget for class '5' for different values of $\lambda$ (Forgetting Lagrangian parameter). Increasing the value of $\lambda$ decreases the remaining information, but increases the error on the remaining samples. The number of samples to forget in the plot varies between one sample and the whole class (404 samples).
    } 
    \label{fig:info-vs-samples}
\end{figure*}

\begin{figure*}[t]
    \centering
    \includegraphics[width=.9\linewidth]{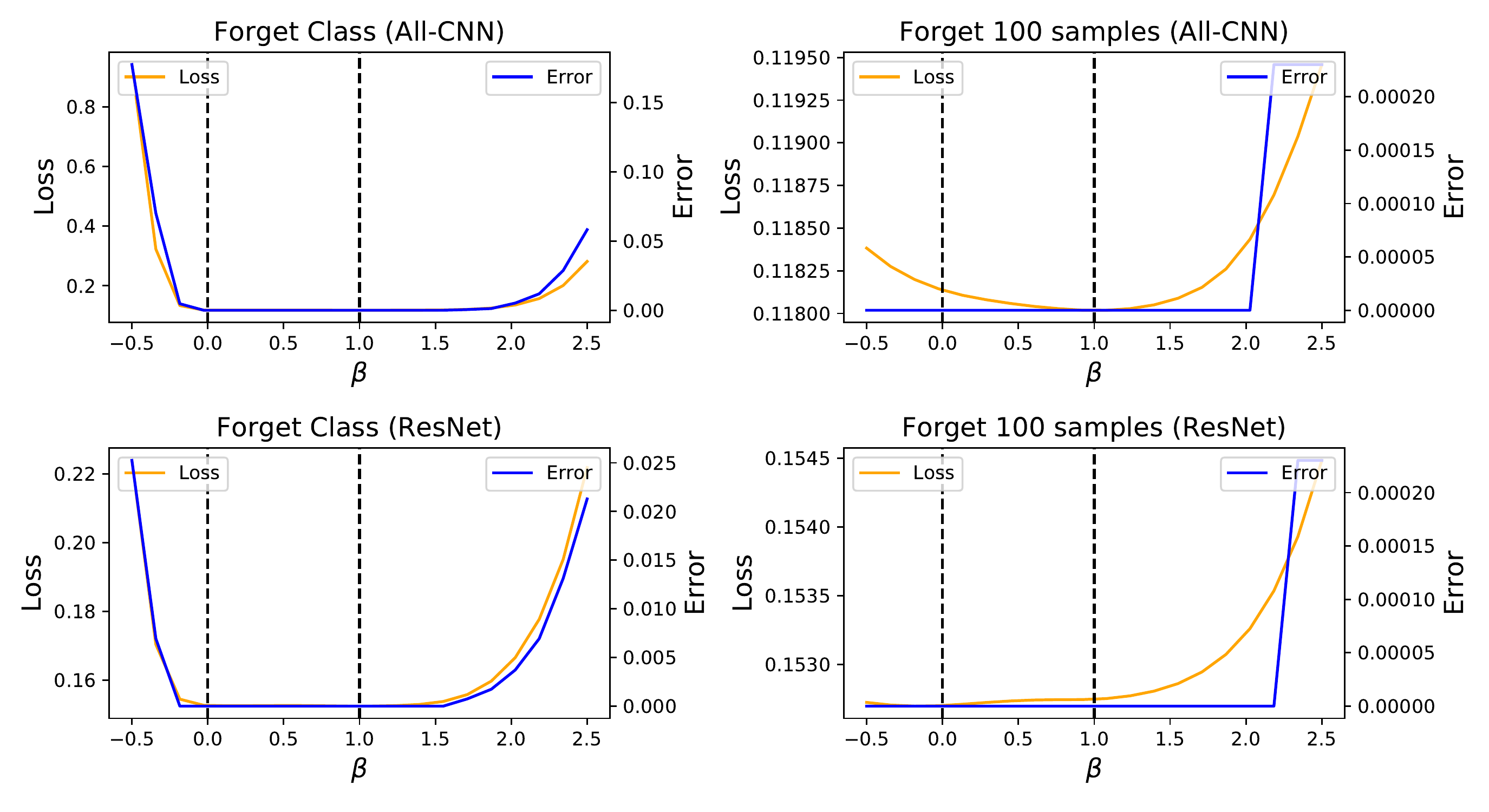}
    \caption{\textbf{Loss landscape of $L_{\Dr}$.} Plot of loss and error on the dataset $\D_r$ of the remaining samples, interpolating along the line joining the \textit{original model} (at $t=0$) and the (target) \textit{retrained model} at $t=1$. Precisely, let $w_o$ be the original model and let $w_r$ be the retrained model, we plot $L(w(t))$, where $w(t)=(1-t) \cdot w_o + t \cdot w_r$ by varying $t \in [-0.5,2.5]$. We observe that the loss along the line joining the original and the target model is convex, and almost flat throughout. In particular, there exists at least an optimal direction (the one joining the two models) along which we can add noise without increasing the loss on the remaining data, and which would allow to forget the extra information. This inspires and justifies our forgetting procedure.
    } 
    \label{fig:loss-landscape}
\end{figure*}

\section{Additional experiments}

In \Cref{fig:loss-landscape} we plot the loss function along a line joining the original model trained on all data and the ground-truth optimal model that was trained from the beginning without seeing the data to forget. The plots confirm that the two models are close to each other, as expected from the stability of SGD, and that the loss function in a neighborhood of the two points is convex (at least on the line joining the two). This justifies the hypotheses that we made to apply our forgetting method (derived in the case of a convex loss) to the more challenging case of deep networks.

\begin{figure*}[h]
    \centering
    \includegraphics[width=1\linewidth]{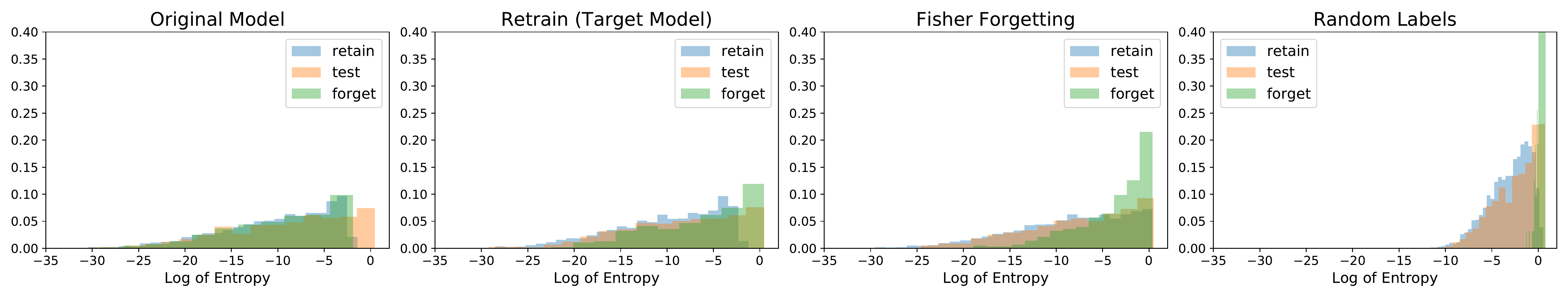}
    \caption{\textbf{Same plot as in \Cref{fig:streisand} but with a different architecture: Streisand Effect for ResNet model}}

    \label{fig:streisand-resnet}
\end{figure*}

\begin{figure*}[h]
    \centering
    \includegraphics[width=.8\linewidth]{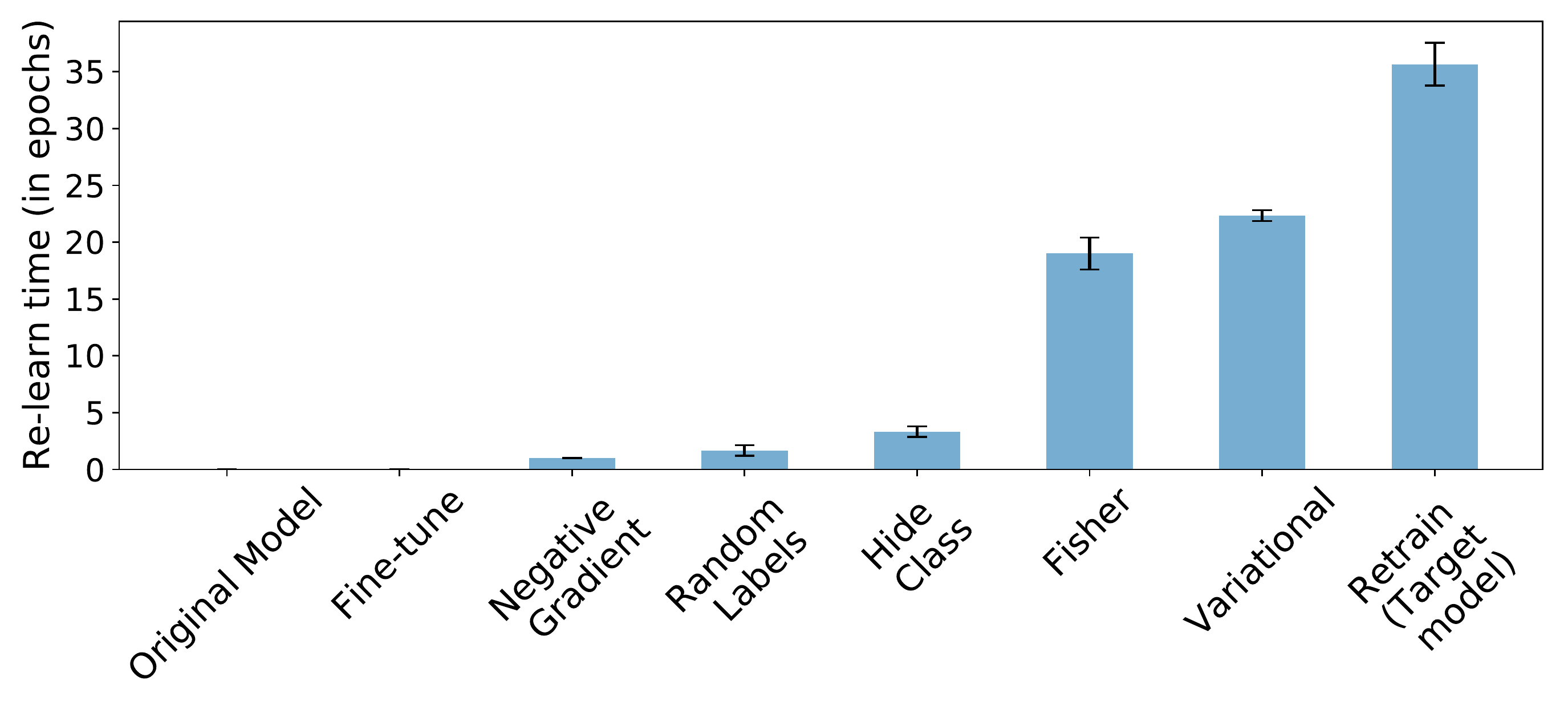}
    \caption{\textbf{ Same plot as in \Cref{fig:lacuna-relearn} but with a different architecture: Re-learn time (in epochs) for various forgetting methods using ResNet model.}
    } 
    \label{fig:lacuna-relearn-resnet}
\end{figure*}

To further confirm that our method can be applied to different architectures, in addition to the experiments on All-CNN that we show in the main paper we run additional experiments on a ResNet-18 architecture. In particular, \Cref{table:results-resnet} show the errors obtained by various forgetting techniques and \Cref{fig:lacuna-relearn-resnet} shows the relearn time for a class after forgetting. In \Cref{fig:streisand-resnet},  we show that ResNets too suffer from a Streisand effect if an improper forgetting procedure is applied.

\section{Implementation details}

The term $e^{-Bt}e^{At}$ in \cref{eq:scrubbing-generic} may diverge for $t \to \infty$ if the actual dynamics of the optimization algorithm do not exactly satisfy the hypotheses. In practice, we found it useful to replace it with $e^{\operatorname{clamp}(A-B, 1/t) t}$, where $\operatorname{clamp}(A, 1/t)$ clamps the eigenvalues of $A$ so that they are smaller or equal to $1/t$, so that the expression does not diverge.%
\footnote{If $A=SDS^T$ is an eigenvalue decomposition of the symmetric matrix $A$, we define $\operatorname{clamp}(A, m) = S\min(D, m)S^T$.}  Notice also that in general $e^{-Bt}e^{At} \neq e^{(A-B)t}$, unless $A$ and $B$ commute, but we found this to be a good approximation in practice.

For DNNs, we perform experiments on Lacuna-10 and CIFAR-10 using All-CNN and ResNet. For All-CNN, we use the model proposed in \cite{springenberg2014striving} and for ResNet we use the ResNet-18 model; however, we reduce the number of filters by half in each layer and remove the final residual block. For all the experiments we first pre-train the network on Lacuna-100 (CIFAR-100), and then fine-tune it on Lacuna-10 (CIFAR-10). We do not use data augmentation in any of the experiments. While pre-training the network we use a constant learning rate of 0.1 and train for 30 epochs and while finetuning we use a constant learning rate of 0.01 and train for 50 epochs. In all the experiments we use weight decay regularization with value 0.0005. We use PyTorch to perform all the experiments.

In \Cref{table:results}, `Original' model denotes the case when we train the model on the entire dataset, ${\cal D}$. `Retrain' denotes the case when we train on ${\cal D}_{r}$ (we can do this by simply replacing the corresponding samples from the data loader). `Finetune' is a possible scrubbing procedure when we use the Original model and fine-tune it on ${\cal D}_{r}$ (we can simply use the data loader from the Retrain case for fine-tuning). We run fine-tuning for 10 epochs with a learning rate 0.01 and weight decay 0.0005. `Negative Gradient' denotes the case when we fine-tune on the complete data; however, for the samples belonging to $\Df$ we simply multiply the gradient by $-1$ (i.e. maximizing the cross-entropy loss for the samples to forget). To prevent the loss from diverging, we clamp the maximum loss to chance level. We train for 10 epochs with learning rate 0.01 and weight decay 0.0005. In `Random Labels,' we randomly shuffle the labels for the samples we want to forget. We use the same training hyper-parameters as the previous methods. For `Hiding,' we replace the row corresponding to the class of the samples to be forgotten in the final classifier layer of the DNN with random initialization. Fisher denotes our method where we estimate the Fisher noise to add to the model. The Fisher noise is computed using a positive semi-definite approximation to the actual Fisher Information Matrix. We compute the trace of the Fisher Information which can be obtained by computing the expected outer product of the gradients of the DNN. In the experiments we observe that $F^{-\frac{1}{2}}$ approximates the Fisher noise  better compared to $F^{-\frac{1}{4}}$. In Variational, we compute the optimal noise to be added for scrubbing a cohort by solving a variational problem. For Fisher and Variational forgetting we choose $\lambda=5 \cdot 10^{-7}$.

In order to compute the Information bound (Fisher and Variational forgetting) we use the Local Forgetting Bound i.e. we apply the scrubbing procedure to both the original and the retrain (target) model and then compute the KL divergence of the two distributions. We use the same random seed while training both the models and use the same scrubbing procedures for both, that is, $S=S_0$. Fisher and Variational forgetting method essentially consists of adding noise to the model weights, i.e., $S(w)=w+n$ where $n \sim {\cal N}(0, \Sigma)$. Let $w_o$, $\Sigma_o$ and $w_r$, $\Sigma_r$ denote the weights and noise covariance for the original and the target retrained model respectively, then the Information bound is given by the following relation:
\begin{multline*}
\KL{{\cal N}(w_o,\Sigma_o)}{{\cal N}(w_r, \Sigma_r)} =\\
\frac{1}{2} \Big( \tr(\Sigma_r^{-1} \Sigma_o) + (w_r-w_o)^{T} \Sigma_r^{-1} (w_r-w_o) - k + \log \frac{|\Sigma_r|}{|\Sigma_o|} \Big),
\end{multline*}
where $k$ is the dimension of $w$. This bound should be computed for multiple values of the initial random seed and then averaged to obtain the Local Forgetting Bound in \Cref{prop:local-forgetting-bound}. In our experiments we compute the bound using a single random seed.

To compute the Relearn-time we train the scrubbed model on the dataset $\D$ for 50 epochs using a constant learning rate 0.01. We report the first epoch when the loss value falls below a certain threshold as the relearn-time.

\subsection{Pre-training improves the forgetting bound}

Our local forgetting bound assumes stability of the algorithm,  which may not be guaranteed when training  a large deep network over many epochs. This can be obviated by pre-training the network, so all paths will start from a common configuration of weights and training time is decreased, and with it the opportunity for paths to diverge. As we show next, the resulting bound is greatly improved. The drawback is that the bound cannot guarantee forgetting of any information contained in the pre-training dataset (that is, $\Df$ needs to be disjoint from $\D_\text{pretrain}$).

\subsection{Datasets}

We report experiments on MNIST, CIFAR10 \cite{krizhevsky2009learning}, Lacuna-10 and Lacuna-100 which we introduce. \textbf{Lacuna-10} consists of face images of 10 celebrities from VGGFaces2 \cite{Cao18}, randomly sampled with at least 500 images each. We split the data into a test set of 100 samples for each class, while the remaining form the training set.  Similarly, \textbf{Lacuna-100} randomly samples 100 celebrities with at least 500 images each. We resize the images to 32x32. There is no overlap between the two Lacuna datasets. We use Lacuna-100 to pre-train (and hence assume that we do not have to forget it), and fine-tune on Lacuna-10. The scrubbing procedure is required to forget some or all images for one identity in Lacuna-10. On both CIFAR-10 and Lacuna-10 we choose to forget either the entire class `5,' which is chosen at random, or a hundred images of the class.

\section{Proofs}

\paragraph{Proposition 1}
Let the forgetting set $\Df$ be a random variable, for instance, a random sampling of the data to forget. Let $Y$ be an attribute of interest that depends on $\Df$. Then,
\begin{align*}
&I(Y; f(S(w))) \leq \nonumber\\ &\quad\quad\E_{D_f}[\KL{P(f(\Scrub(w))|\D)}{P(f(\Scrub_0(w))|\Dr)}].
\end{align*}

\begin{proof}
We have the following Markov Chain:
\[
Y \longleftarrow \D_f \longrightarrow w \longrightarrow S(w) \longrightarrow f(S(w))
\]
We assume that the retain set $\Dr$ is fixed and thus deterministic. Also, $\Df \rightarrow w$ implies that we first sample $\Df$ and then use $\D = \Dr \text{(fixed and known)} \cup \Df$ for training $w$.
From Data Processing Inequality, we know that: $I(Y; f(S(w))) \leq I(\Df; f(S(w)))$.
In order to prove the proposition we bound the
RHS term (i.e $I(\Df; f(S(w)))$). 
\begin{align*}
&= I(\Df; f(S(w))) \\
&= \E_{\Df}\big[\KL{ p(f(S(w)) | \Df \cup \Dr) }{ p(f(S(w)))}\big]\\
&= \E_\Df \E_{p(f(S(w)) | \Df \cup \Dr)}\Big[ \log \frac{p(f(S(w)) | \Df \cup \Dr)}{ p(f(S(w)))}\Big]\\
&= \E_\Df \E_{p(f(S(w)) | \Df \cup \Dr)}\Big[  \log \frac{p(f(S(w)) | \Df \cup \Dr)}{ p(f(S_0(w))|\Dr)}
\\
&\hspace{5cm}+
\log \frac{p(f(S_0(w))|\Dr)}{ p(f(S(w)))}
\Big]\\
&=\E_{\Df}\big[\KL{ p(f(S(w)) | \Df \cup \Dr)}{ p(f(S_0(w)) | \Dr)}\big]\\
&\hspace{3cm}
- \KL{p(f(S_0(w)))}{p(f(S(w)) | \Dr)}\\
&\leq \E_{\Df}\big[\KL{ p(f(S(w)) | \Df \cup \Dr) }{ p(f(S_0(w)) | \Dr)}\big]
\end{align*}
where the last inequality follows from the fact that KL-divergence is always non-negative.

\end{proof}

\paragraph{Lemma 1}
For any function $f(w)$ have
\begin{multline*}
\KL{P(f(\Scrub(w))|\D)}{P(f(\Scrub_0(w))|\Dr)}\\
\leq \KL{P(\Scrub(w)|\D)}{P(\Scrub_0(w)|\Dr)},    
\end{multline*}
\begin{proof}
For simplicity sake, we will consider the random variables to be discrete. To keep the notation uncluttered, we consider the expression
\[\KL{Q(f(x))}{R(f(x))}\\
\leq \KL{Q(x)}{R(x)}\] 
where $Q(w)=P(\Scrub(w)|{\cal D})$ and $R(w)=P(\Scrub_0(w)|{\cal D}_{-k})$.
Let $W_c = \{ w | f(w)=c\}$, so that we have $Q(f(w)=c)=\sum_{w \in W_c} Q(w)$ and similarly $R(f(w)=c)=\sum_{w \in W_c} R(w)$. 
Rewriting the LHS with this notation, we get:
\begin{align}
&\KL{Q(f(w))}{R(f(w))}\nonumber \\
&\quad\quad=\sum_c Q(f(w)=c) \log \frac{Q(f(w) = c)}{R(f(w)=c)} \nonumber\\
&\quad\quad= \sum_c Q(f(w)=c) \log \frac{\sum_{w\in W_c} Q(w)}{\sum_{w\in W_c} R(w)}
\label{eq:lemma-lhs}
\end{align}
We can similarly rewrite the RHS:
\begin{align}
\label{eq:lemma-rhs}
&\KL{Q(w)}{R(w)} = \sum_{c} \sum_{w\in W_c} \log{\frac{Q(w)}{R(w)}}
\end{align}
From the log-sum inequality, we know that for each $c$ in eq. (\ref{eq:lemma-lhs}) and eq.  (\ref{eq:lemma-rhs}):
\begin{multline*}
\big(\sum_{w 
\in W_c} Q(w)\big) \log{\frac{\sum_{w 
\in W_c} Q(w)}{\sum_{w
\in W_c} R(w)}} \leq \\ \sum_{w
\in W_c} Q(w) \log{\frac{Q(w)}{R(w)}}    
\end{multline*}
Summation over all $c$ on both sides of the inequality concludes the proof.
\end{proof}

\paragraph{Proposition 2}
Let $A(\D)$ be a (possibly stochastic) training algorithm, the outcome of which we indicate as $w = A(\D, \epsilon)$ for some deterministic function $A$ and $\epsilon \sim {\cal N}(0, I)$, for instance a random seed. Then, we have $P(S(w)|\D) = \E_\epsilon[P(S(w)|\D, \epsilon)]$. We have the bound:
\begin{multline*}
\KL{P(S(w)|\D)}{P(S_0(w)|\Dr)} \leq\\
 \E_\epsilon\Big[ \KL{P(S(w)|\D, \epsilon)}{P(S_0(w)|\Dr, \epsilon)} \Big]
\end{multline*}
\begin{proof}
To keep the notation uncluttered,we rewrite the inequality as:
\[\KL{Q(w)}{R(w)} \leq\\
 \E_\epsilon\Big[ \KL{Q(w| \epsilon)}{R(w |\epsilon)} \Big]\]
where $Q(w) = P(\Scrub(w)|\D)$ and $R(w) = P(\Scrub(w)|\Dr)$
The LHS can be equivalently written as:
\begin{align*}
&\KL{Q(w)}{R(w)} = \\
&\quad\quad= \int Q(w) \log \frac{Q(w)}{R(w)}) dw\\
&\quad\quad= \int \E_\epsilon[Q(w|\epsilon)]\log \frac{\E_\epsilon[Q(w|\epsilon)]}{\E_\epsilon[P(w|\epsilon)]} dw\\
&\quad\quad\stackrel{(*)}{\leq}\int \E_\epsilon \Big[ Q(w|\epsilon) \log \frac{Q(w|\epsilon)}{P(w|\epsilon)}\Big] dw\\
&\quad\quad=\E_\epsilon \Big[ \int Q(w|\epsilon) \log \frac{Q(w|\epsilon)}{P(w|\epsilon)} dw \Big]\\
&\quad\quad= \E_\epsilon \big[\KL{Q(w|\epsilon)}{R(w|\epsilon)}\big],
\end{align*}
where in (*) we used the log-sum inequality. 
\end{proof}

\paragraph{Proposition 3}
Let the loss be $L_{\D}(w) = L_{\Df}(w) + L_{\Dr}(w)$, and assume both $L_\D(w)$ and $L_\Dr(w)$ are quadratic. Assume that the optimization algorithm $A_t(\D, \epsilon)$ at time $t$ is given by the gradient flow of the loss with a random initialization, and let $h(w)$ be the function:
\[
h(w) = e^{-Bt}e^{At} d+ e^{-Bt}(d-d_r) - d_r,
\]
where $A = \nabla^2 L_\D(w)$, $B = \nabla^2 L_\Dr(w)$, $d = A^{-1} \nabla_w L_\D$ and $d_r = B^{-1} \nabla_w L_\Dr$, and $e^{At}$ denotes the matrix exponential. Then $h(A_t(\D, \epsilon)) = A_t(\D_r, \epsilon)$ for all random initializations $\epsilon$ and all times $t$. 

\begin{proof}
Since $L_\D$ and $L_\Dr$ are quadratic, assume without loss of generality that:
\begin{align*}
L_\D(w) &= \frac{1}{2}(w - w^*_A)^T A (w - w^*_A) \\
L_\Df(w) &= \frac{1}{2}(w - w^*_B)^T B (w - w^*_B)
\end{align*}
Since the training dynamic is given by a gradient flow, the training path is the solution to the differential equation:
\begin{align*}
\dot{w_A}(t) &= A (w(t) - w^*_A) \\
\dot{w_B}(t) &= B (w(t) - w^*_B) 
\end{align*}
which is given respectively by:
\begin{align*}
w_A(t) &= w^*_A + e^{-At} (w_0 - w^*_A) \\
w_B(t) &= w^*_B + e^{-Bt} (w_0 - w^*_B) 
\end{align*}
We can compute $w_0$ from the first expression:
\[
w_0 = e^{At} (w_A(t) - w^*_A) + w^*_A = e^{At} d_A + w^*_A,
\]
where we defined $d_A = w_A(t) - w^*_A = A^{-1} \nabla_w L_\D(w_A(t))$. We now replace this expression of $w_0$ in the second expression to obtain:
\begin{align*}
w_B(t) &=  e^{-Bt}e^{At} d_A + e^{-Bt} (w^*_A - w^*_B) + w^*_B \\
&= w_A(t) + e^{-Bt}e^{At} d_A + e^{-Bt} (d_B - d_A) - d_B
\end{align*}
where $d_B := w_A(t) - w^*_B = B^{-1} \nabla_w L_\Dr(w_A(t))$.
\end{proof}

\paragraph{Proposition 4}

Assume that $h(w)$ is close to $w'$ up to some normally distributed error $h(w) - w' \sim N(0, \Sigma_h)$, and assume that $L_\Dr(w)$ is (locally) quadratic around $h(w)$. Then the optimal scrubbing procedure
in the form
$%
S(w) = h(w) + n, \quad n \sim N(0, \Sigma), 
 $%
that minimizes the Forgetting Lagrangian
\begin{multline*}
\mathcal{L} = \E_{\tilde{w} \sim S(w)}[L_\Dr(\tilde{w})] + \\\lambda\, 
 \E_\epsilon\Big[ \KL{P(S(w)|\D, \epsilon)}{P(S_0(w)|\Dr, \epsilon)} \Big]
\end{multline*}
is obtained when $\Sigma B \Sigma = \lambda \Sigma_h$, where $B = \nabla^2 L_\Dr(k)$. In particular, if the error is isotropic, that is $\Sigma_h = \sigma_h^2 I$ is a multiple of the identity, we have $\Sigma = \sqrt{\lambda \sigma_h^2} B^{-1/2}$.

\begin{proof}
We consider the following second order approximation to the loss function in the neighbourhood of the parameters at convergence:
\begin{align*}
&\E_{\tilde{w} \sim S(w)}[L_\Dr(\tilde{w})] =\E_{n \sim N(0,\Sigma)}[L_\Dr(h(w) + n)]\\
&\quad=\E_{n \sim N(0,\Sigma)} \Big[ L_{{\cal D}_{r}}(h(w))\ +\\ 
&\quad\quad\quad\quad\nabla L_{{\cal D}_{r}} (h(w))^{T} \Sigma^{\frac{1}{2}} n + \frac{1}{2}(\Sigma^{\frac{1}{2}}n)^{T}B(\Sigma^{\frac{1}{2}}n) + o(n^2) \Big]\\
&\quad\simeq L_{\Dr}(h(w)) + \frac{1}{2} \E_{n \sim N(0,\Sigma)} \Big[ (\Sigma^{\frac{1}{2}}n)^{T}B(\Sigma^{\frac{1}{2}}n) \Big]\\
&\quad= L_{\Dr}(h(w)) + \frac{1}{2} \E_{n \sim N(0,\Sigma)} \Big[ \operatorname{tr}( B(\Sigma^{\frac{1}{2}}n n^{T} \Sigma^{\frac{1}{2}})) \Big]\\
&\quad= L_{\Dr}(h(w)) +  \frac{1}{2} \tr(B\Sigma)
\end{align*}

Recall that we take $S_0$ to be $S_0(w) = w + n'$ with $n' \sim N(0, \Sigma)$, so that $P(S_0(w)|\Dr, \epsilon) \sim N(w, \Sigma)$. Thus, we get the following expression for the KL divergence term: 
\begin{align*}
&\E_\epsilon\Big[ \KL{P(S(w)|\D, \epsilon)}{P(S_0(w)|\Dr, \epsilon)} \Big]\\
&\quad= \frac{1}{2} \E_{\epsilon} \Big[ (h(w) - w')^T\Sigma^{-1}(h(w) - h(w')) \Big]\\
&\quad=\frac{1}{2}  \tr(\Sigma^{-1} \Sigma_{h}),
\end{align*}
where in the last equality we have used that by hypothesis $h(w) - w' \sim N(0, \Sigma_h)$.
Combining the two terms we get: 
\begin{align*}
\mathcal{L} = L_{\Dr}(h(w)) +  \frac{1}{2} \tr(B\Sigma) + \frac{1}{2}  \tr(\Sigma^{-1} \Sigma_{h})
\end{align*}
We now want to find the optimal covariance $\Sigma$ of the noise to add in order to forget. Setting $\nabla_{\Sigma} \mathcal{L} = 0$, we obtain the following optimality condition $\Sigma B \Sigma =  \lambda \Sigma_{h}$. If we further assume the error $\Sigma_h$ to be isotropic, that is, $\Sigma_h = \sigma_h^2 I$, then this condition simplifies to $\Sigma = \sqrt{\lambda \sigma_h^2} B^{-1/2}$.
\end{proof}

\begin{table*}[h]
\centering
\caption{
\label{table:results-resnet}
\textbf{Same experiment as \Cref{table:results} but with a different architecture: Error readout functions for ResNet trained on Lacuna-10 and CIFAR-10.}
}
\vspace{.5em}
\newcommand{\std}[1]{\color{black!70}{$\pm$#1}}
\resizebox{2\columnwidth}{!}{
\begin{adjustbox}{center}
{\small
\begin{tabular}{p{0.075\linewidth}|p{0.14\linewidth}|p{0.075\linewidth}|p{0.075\linewidth}|p{0.074\linewidth}p{0.073\linewidth}p{0.074\linewidth}p{0.074\linewidth}|p{0.085\linewidth}p{0.074\linewidth}}
\hline
  &  Metrics & \small{\textbf{Original model}} & \small{\textbf{Retrain} (target)} & \small{\textbf{Finetune}} & \small{\textbf{Neg. Grad.}} & \small{\textbf{Rand. Lbls.}}  & \small{\textbf{Hiding}}  & \small{\textbf{Fisher} (ours)} & \small{\textbf{Variational} (ours)}  \\
\hline
\footnotesize{Lacuna-10} & Error on $\mathcal{D}_\text{test}$ \scriptsize{(\%)} & 
\footnotesize{18.0} \std{0.5} & 
\footnotesize{18.2} \std{0.3} & 
\footnotesize{18.1} \std{0.2} & 
\footnotesize{18.0} \std{0.5}  & 
\footnotesize{19.3} \std{0.7} & 
\footnotesize{25.1} \std{0.1} & 
\footnotesize{24.5} \std{0.7} & 
\footnotesize{28.2} \std{3.2}\\
\scriptsize{Scrub 100} & Error on $\mathcal{D}_f$ \scriptsize{(\%)} & 
\footnotesize{0.0} \std{0.0} & 
\footnotesize{29.0} \std{0.1} & 
\footnotesize{0.0} \std{0.0} & 
\footnotesize{0.0} \std{0.0} & 
\footnotesize{14.3} \std{6.7} & 
\footnotesize{100} \std{0.0} & 
\footnotesize{17.7} \std{6.7} & 
\footnotesize{3.0} \std{1.0}\\
\scriptsize{ResNet} & Error on  $\mathcal{D}_r$ \scriptsize{(\%)} & 
\footnotesize{0.0} \std{0.0} & 
\footnotesize{0.0} \std{0.0} & 
\footnotesize{0.0} \std{0.0} & 
\footnotesize{0.0} \std{0.0} & 
\footnotesize{0.2} \std{0.0} & 
\footnotesize{6.5} \std{0.0} & 
\footnotesize{10.4} \std{0.8} & 
\footnotesize{11.7} \std{5.8}\\
\scriptsize{} & Info-bound \scriptsize{(kNATs)} &    &  &  &   &  &  & \footnotesize{2.6} \std{0.1} &  \footnotesize{2.6} \std{0.3}\\
\hline
\footnotesize{Lacuna-10} & Error on $\mathcal{D}_\text{test}$ \scriptsize{(\%)} & 
\footnotesize{18.0} \std{0.5} & 
\footnotesize{24.5} \std{0.4} & 
\footnotesize{18.1} \std{0.4} & 
\footnotesize{25.0} \std{0.4}  & 
\footnotesize{24.6} \std{0.8} & 
\footnotesize{22.0} \std{5.2} & 
\footnotesize{26.6} \std{1.0} & 
\footnotesize{26.8} \std{0.6}\\
\scriptsize{Forget class} & Error on $\mathcal{D}_f$ \scriptsize{(\%)} & 
\footnotesize{0.0} \std{0.0} & 
\footnotesize{100} \std{0.0} & 
\footnotesize{0.0} \std{0.0} & 
\footnotesize{99.7} \std{0.3}  & 
\footnotesize{90.3} \std{1.5} & 
\footnotesize{100.0} \std{0.0} & 
\footnotesize{100.0} \std{0.0} & 
\footnotesize{100.0} \std{0.0}\\

\scriptsize{ResNet} & Error on  $\mathcal{D}_r$ \scriptsize{(\%)} & 
\footnotesize{0.0} \std{0.0} & 
\footnotesize{0.0} \std{0.0} & 
\footnotesize{0.0} \std{0.0} & 
\footnotesize{0.0} \std{0.0}  & 
\footnotesize{0.0} \std{0.0} & 
\footnotesize{0.0} \std{0.0} & 
\footnotesize{0.2} \std{0.4} & 
\footnotesize{1.0} \std{0.9}\\
& Info-bound \scriptsize{(kNATs)}&  & &  & &  &  &  
\footnotesize{33.2} \std{2.2} &  \footnotesize{25.7} \std{7.8}\\
\hline
\footnotesize{CIFAR-10} & Error on $\mathcal{D}_\text{test}$ \scriptsize{(\%)} & 
\footnotesize{17.3} \std{0.2} & 
\footnotesize{17.5} \std{1.0} & 
\footnotesize{17.3} \std{0.5} & 
\footnotesize{17.2} \std{0.3} & 
\footnotesize{17.9} \std{1.2} & 
\footnotesize{23.6} \std{1.3} & 
\footnotesize{21.9} \std{2.2} & 
\footnotesize{23.1} \std{2.3}\\
\scriptsize{Scrub 100} & Error on $\mathcal{D}_f$ \scriptsize(\%)& 
\footnotesize{0.0} \std{0.0} & 
\footnotesize{25.2} \std{5.2} & 
\footnotesize{0.0} \std{0.0} & 
\footnotesize{0.0} \std{0.0} & 
\footnotesize{0.0} \std{0.0} & 
\footnotesize{100.0} \std{0.0} & 
\footnotesize{16.0} \std{7.2} & 
\footnotesize{13.3} \std{7.4}\\
\scriptsize{ResNet} & Error on  $\mathcal{D}_r$ \scriptsize(\%)& 
\footnotesize{0.0} \std{0.0} & 
\footnotesize{0.0} \std{0.0} & 
\footnotesize{0.0} \std{0.0} & 
\footnotesize{0.0} \std{0.0} & 
\footnotesize{0.0} \std{0.0} & 
\footnotesize{8.7} \std{1.9} & 
\footnotesize{7.6} \std{4.0} & 
\footnotesize{9.8} \std{3.9}\\
\scriptsize{} & Info-bound \scriptsize{(kNATs)} &  &  &  &   &  &  & \footnotesize{46.0} \std{19.3} &  \footnotesize{29.8} \std{8.0}\\
\hline
\footnotesize{CIFAR-10} & Error on $\mathcal{D}_\text{test}$ \scriptsize{(\%)} & 
\footnotesize{17.1} \std{0.3} & 
\footnotesize{22.8} \std{0.0} & 
\footnotesize{17.2} \std{0.1} & 
\footnotesize{22.8} \std{0.1} & 
\footnotesize{23.1} \std{0.2} & 
\footnotesize{22.8} \std{0.1} & 
\footnotesize{25.8} \std{0.1} & 
\footnotesize{25.1} \std{0.3}\\
\scriptsize{Forget class} & Error on $\mathcal{D}_f$ \scriptsize{(\%)} & 
\footnotesize{0.0} \std{0.0} & 
\footnotesize{100} \std{0.0} & 
\footnotesize{0.7} \std{0.5} & 
\footnotesize{100} \std{0.1}  & 
\footnotesize{94.2} \std{5.7} & 
\footnotesize{100.0} \std{0.0} & 
\footnotesize{100.0} \std{0.0} & 
\footnotesize{100.0} \std{0.0}\\
\scriptsize{ResNet} & Error on  $\mathcal{D}_r$ \scriptsize{(\%)} & 
\footnotesize{0.0} \std{0.0} & 
\footnotesize{0.0} \std{0.0} & 
\footnotesize{0.0} \std{0.0} & 
\footnotesize{0.0} \std{0.0}  & 
\footnotesize{0.0} \std{0.0} & 
\footnotesize{0.0} \std{0.0} & 
\footnotesize{3.8} \std{0.4} & 
\footnotesize{2.4} \std{1.0}\\
& Info-bound \scriptsize{(kNATs)}& & &  & &  &  &  
\footnotesize{235.4} \std{4.9} &  \footnotesize{234.8} \std{6.1}\\
\hline
\hline
\end{tabular}
}
\end{adjustbox}
}

\end{table*}

\end{document}